\documentclass[letterpaper, 10 pt, conference]{ieeeconf}  

\IEEEoverridecommandlockouts                              

\usepackage{graphics} 
\usepackage{mathptmx} 
\usepackage{amsmath} 
\usepackage{amssymb}  
\usepackage{booktabs} 
\usepackage{colortbl} 
\usepackage{xcolor} 
\usepackage{lipsum} 

\usepackage{pifont} 
\usepackage{cuted}

\usepackage[most]{tcolorbox}
\usepackage{afterpage}
\tcbuselibrary{listings}
\usepackage{caption}
\usepackage{url}
\usepackage{hyperref}

\usepackage{multirow}

\usepackage[noadjust]{cite}

\usepackage{lipsum} 
\usepackage{tikz}
\usepackage{atbegshi} 

\title{\LARGE \bf
This\&That: Language-Gesture Controlled \\Video Generation for Robot Planning
}

\author{%
  Boyang Wang$^1$\quad Nikhil Sridhar$^1$\quad Chao Feng$^1$\quad Mark Van der Merwe$^1$ \\
  Adam Fishman$^2$\quad Nima Fazeli$^1$\quad Jeong Joon Park$^1$\\
  $^1$University of Michigan \quad
  $^2$University of Washington
}

\begin{document}

\maketitle

\begin{strip}
\vspace{-1.8cm}
\begin{center}
    \includegraphics[width=0.95\textwidth]{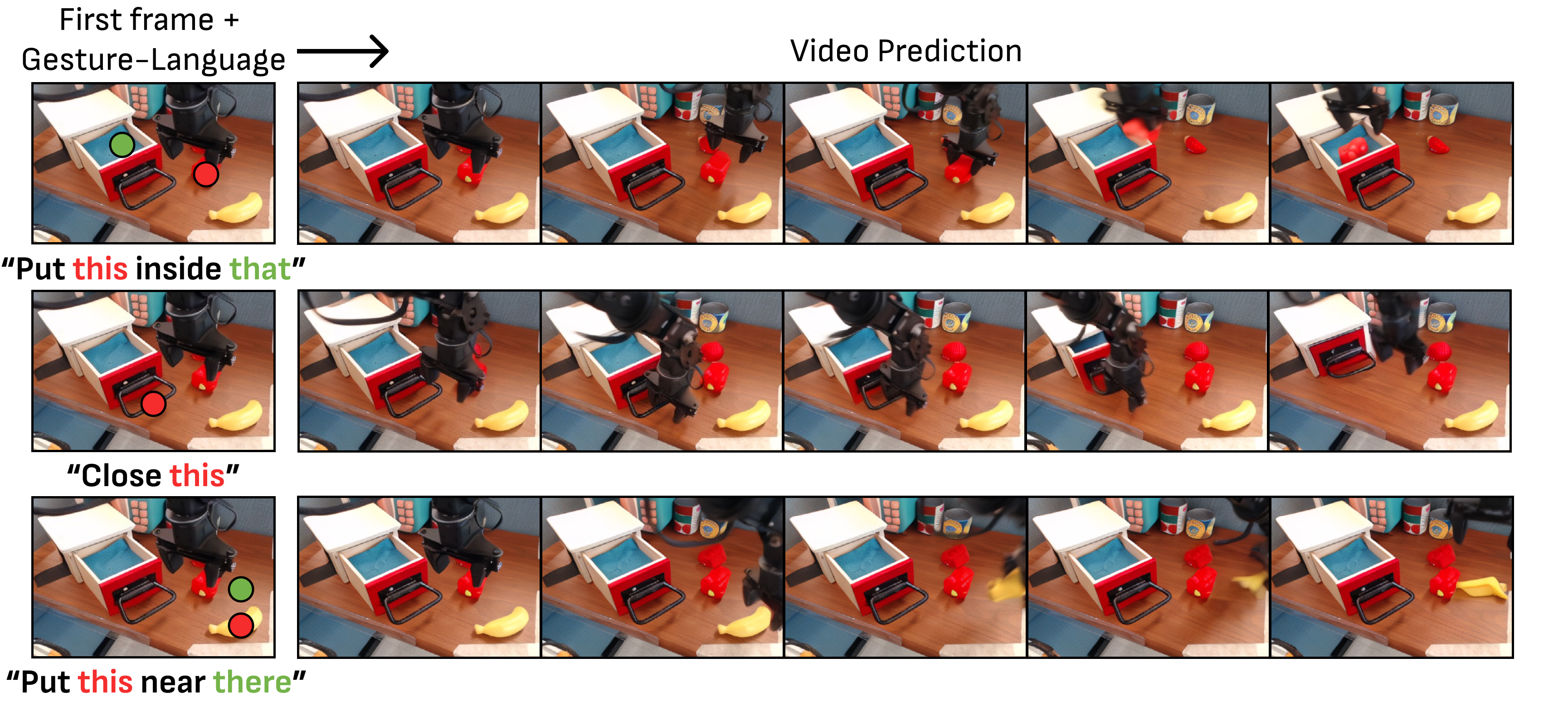}

    \captionof{figure}{\textbf{Video generation for robot planning.} Using the same initial frame, our video diffusion model can effectively generate various action sequences, each conditioned on different pairs of gestures and text prompts. Our approach accommodates simple deictic language such as {\em this} and {\em that}. Our gesture conditioning proves critical for precise video control.}
  \label{fig:teaser}
\end{center}
  \vspace{-0.5cm}
\end{strip}

\begin{abstract}
Clear, interpretable instructions are invaluable when attempting any complex task. Good instructions help to clarify the task and even anticipate the steps needed to solve it. In this work, we propose a robot learning framework for communicating, planning, and executing a wide range of tasks, dubbed {\em This\&That}.
{\em This\&That} solves general tasks by leveraging video generative models, which, through training on internet-scale data, contain rich physical and semantic context. In this work, we tackle three fundamental challenges in video-based planning: 1) unambiguous task communication with simple human instructions, 2) controllable video generation that respects user intent, and 3) translating visual plans into robot actions. {\em This\&That} uses language-gesture conditioning to generate video predictions, as a succinct and unambiguous alternative to existing language-only methods, especially in complex and uncertain environments. These video predictions are then fed into a behavior cloning architecture dubbed Diffusion Video to Action (DiVA), which outperforms prior state-of-the-art behavior cloning and video-based planning methods by substantial margins. 
Project website: \href{https://this-and-that-vid.github.io/this-and-that/}{https://this-and-that-vid.github.io/this-and-that/}.
\end{abstract}

\section{Introduction}
\label{sec:intro}
When we instruct other people to perform a task, we often point to the targets and say things like: \textit{``Give me that glass''} or \textit{``Put this there.''} Such simple language-gesture instructions can be more effective in communicating tasks than verbally describing them without gestures. For example, the verbal instruction -- {\em ``Give me the blue glass located on the third row of the wooden cabinet''} -- can be verbose and ambiguous. The combination of pointing gestures and {\em diectic} words such as ``this" and ``that" is convenient and clear at the same time, so it is used widely across cultures. 
It would be practical if we could control robots on a wide range of tasks using these simple language-gesture commands. Through this work, we strive to achieve exactly {\em that}. 

Our proposed framework, dubbed {\em This\&That}, includes a controllable video generation module and a video-driven robot execution module. We build our video generator on top of a recent large-scale, open-vocabulary text-to-video diffusion model \cite{blattmann2023stable}, which we fine-tune on robotics scenes. Our video diffusion model (VDM) is conditioned on diectic language describing the task as well as gestures represented as 2D locations in the first frame image. We introduce novel techniques to incorporate the multi-modal conditionings, leading to SOTA-quality videos that closely align with human intent, even for uncertain tasks in complex scenes. 

Inspired by recent developments, we consider the video generator as a generalizable planner that envisions how the environment changes for a wide range of tasks. Here, the predicted video is a guide for robot actions, and the execution module only has to follow the predicted video. Unlike existing video-based approaches that either simplify the action space \cite{ko2023learning} or devise special inverse dynamics models \cite{yang2023learning}, we propose a novel integration of video prediction into the broader framework of behavior cloning. Our BC-based execution module efficiently cross-attends to the video frames to unify video-based planning and manipulation. 

We conduct experiments on the Bridge video datasets \cite{walke2023bridgedata,ebert2021bridge} and in IsaacGym simulation. Our experiments include a wide range of open-vocabulary single-step manipulation tasks within complex and uncertain environments. The results demonstrate that our {\em This\&That} framework produces higher quality videos with superior alignment to user intent than prior work. Our behavior cloning experiments in simulated environments show the benefits of our proposed language-gesture commands and further justify the use of video planning for policy learning.
Overall, we claim the following two contributions: 

\begin{itemize}
    \item We propose a novel video generation method that combines language and gesture instruction to create robot action plans that more accurately reflect user intent compared to previous language-only approaches.
    \item We devise a video-conditioned behavior cloning architecture to integrate the generated video predictions with live observations for multi-task robot policy learning.
\end{itemize}

\section{Related Work}
\label{sec:related_work}

\noindent\textbf{Imitation Learning.} Imitation learning is a technique to learn robotic behavior from expert demonstration. Behavior cloning (BC) casts this task as supervised learning by training a policy to directly mimic the expert's actions. Recently, BC has demonstrated strong performance in learning complex, dextrous robot skills~\cite{shridhar2022perceiveractor, chi2023diffusionpolicy, zhao2023learning, brohan2023rt1, florence2021implicit}. While some BC policies rely solely on the robot's state and visual observation, adding goal information to the policy allows the robot to disambiguate between tasks~\cite{gcsup}. Goals can be expressed as language instruction~\cite{brohan2023rt1, lynch2022interactive, karamcheti2023languagedriven, lynch2021language, shao2020concept}, images~\cite{Danielczuk_2019}, or sketches~\cite{sundaresan2024rtsketch}. When visual goals are expressed as a single image, as in \cite{jang2022bcz, brohan2023rt1, burns2022robust}, the goals can be ambiguous, and models tend to overfit. Instead, we represent goals as a dense sequence of images, predicted by a video diffusion model. We find this formulation better aids the policy in reaching long-term goals. 

\noindent\textbf{Video Diffusion Models.} A Video Diffusion Model (VDM) aims to generate temporally consistent and high-fidelity videos~\cite{singer2022make, menapace2024snap, ren2024consisti2v} that align with provided conditions, which may include image~\cite{zhou2024storydiffusion}, text~\cite{henschel2024streamingt2v}, audio~\cite{tian2024emo}, segmentation~\cite{wu2024draganything, ma2024follow, guo2023sparsectrl}, camera pose~\cite{he2024cameractrl, kuang2024collaborative}, human pose~\cite{zhu2024champ, hu2023animate}, etc. 
In VDM literature, significant attention has been given to motion and camera control. MotionCtrl~\cite{wang2023motionctrl} proposes independent modules to control object motion and camera motion by fusing object trajectory and camera pose information into the VDM's convolutional and temporal layers respectively. DragAnything~\cite{wu2024draganything} and Follow-Your-Click~\cite{ma2024follow} leverage segmentation masks to identify target objects and provide complete trajectory data to each frame during training, thus controlling object motion in generated videos. SparseCtrl~\cite{guo2023sparsectrl} introduces channel-wise concatenation to address the challenges posed by sparse temporal conditioning during inference. However, these methods typically require either dense spatial or dense temporal information, or both, for effective motion control. In contrast, our approach requires input as sparse as two points, with all other information inferred by the model based on learned statistical patterns.

\noindent\textbf{Video Diffusion Models in Robotics.} 
Considerable progress has been made in advancing VDM for use in downstream robotics application~\cite{du2024learning, zhou2024robodreamer, du2023video, yang2023learning, ko2023learning, black2023zero}, yet both the mechanisms of directed video prediction and video-conditioned control are still under-explored. UniPi~\cite{du2023video} and UniSim~\cite{yang2023learning} demonstrate how basic image and text-to-video generation can simulate robotic interactions in real-world scenes. In these works, generated videos are converted into robot actions through inverse dynamics modeling. SuSIE~\cite{black2023zero} employs an autoregressive image editing diffusion model to predict the next visual sub-goal, on which the learned low-level policy is conditioned on.
AVDC~\cite{ko2023learning} proposes a custom VDM architecture to generate low-resolution video plans. It extracts optical flow information from generated video to convert frames into discrete actions. However, these formulations cannot accurately predict behavior in ambiguous scenes, a problem we address by conditioning video generation on both language and \textit{gesture}.

\begin{figure}
    \centering
    \includegraphics[width=1.0\columnwidth]{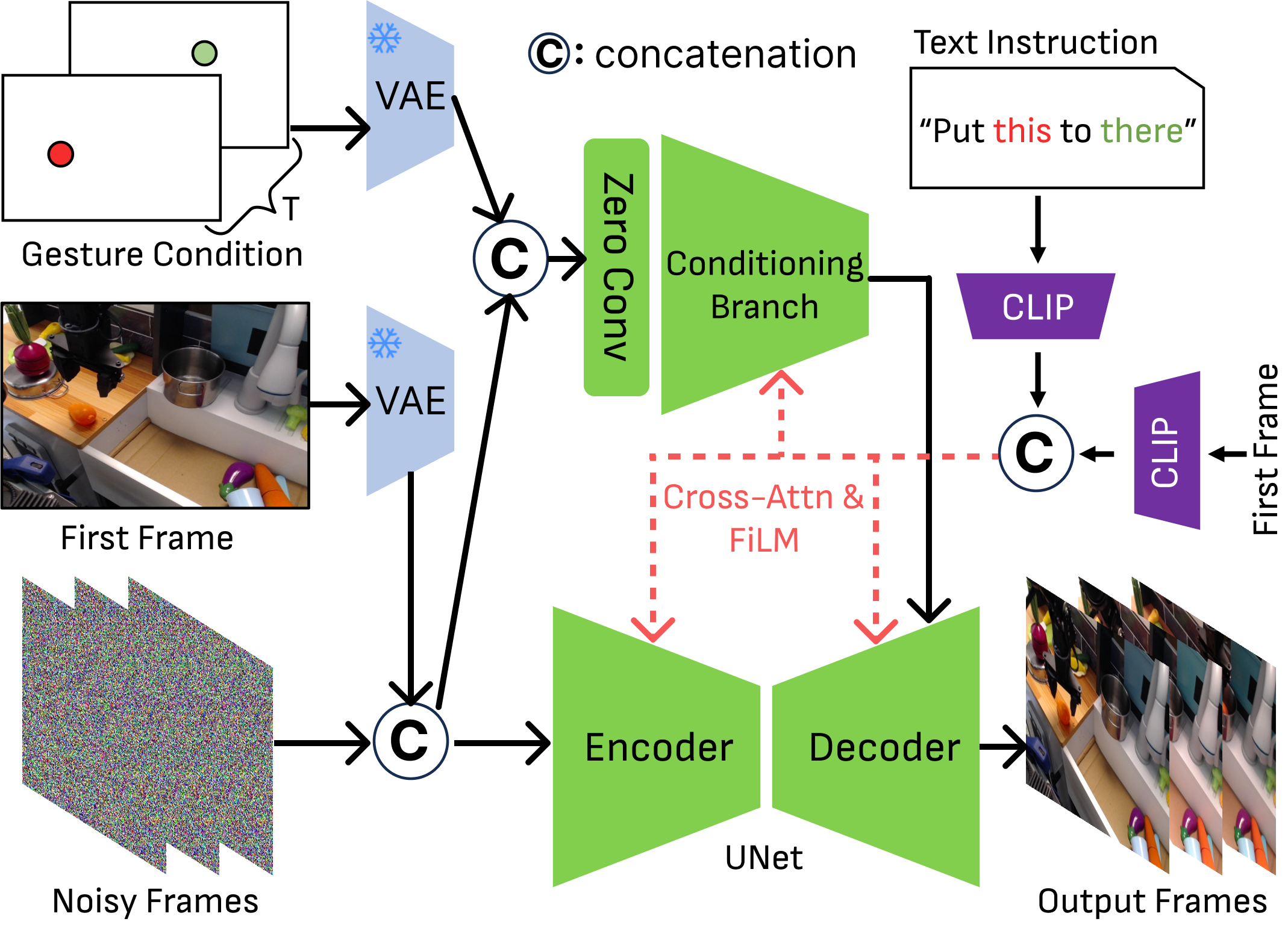}
    \caption{\textbf{Video Diffusion Model Architecture.} Our video diffusion model architecture with first frame image and language-gesture conditioning.}
    \label{fig:diffusion}
    \vspace{-13pt}  
\end{figure}

\section{Overview}

\vspace{-10pt}

Our proposed {\em This\&That} framework is composed of two components: language-gesture conditioned video generation and video-based robot planning. In Sec.~\ref{subsec:videodiffusion}, we introduce our video diffusion model built on top of a large video model. Notably, we use language-gesture conditioning to provide user-friendly control (Sec~\ref{subsec:this_that_cond}). In Sec.~\ref{subsec:videotorobot}, we introduce our behavioral cloning approach which learns to convert generated video plans into low-level robot actions. In the experiment section (Sec.~\ref{sec:experiment}), we prove that videos generated from our language-gesture conditioned diffusion model align better with user intent than prior work and can be used downstream to solve tasks in simulated environments.

\begin{figure*}[t]
    \centering
    \includegraphics[width=0.875\textwidth]{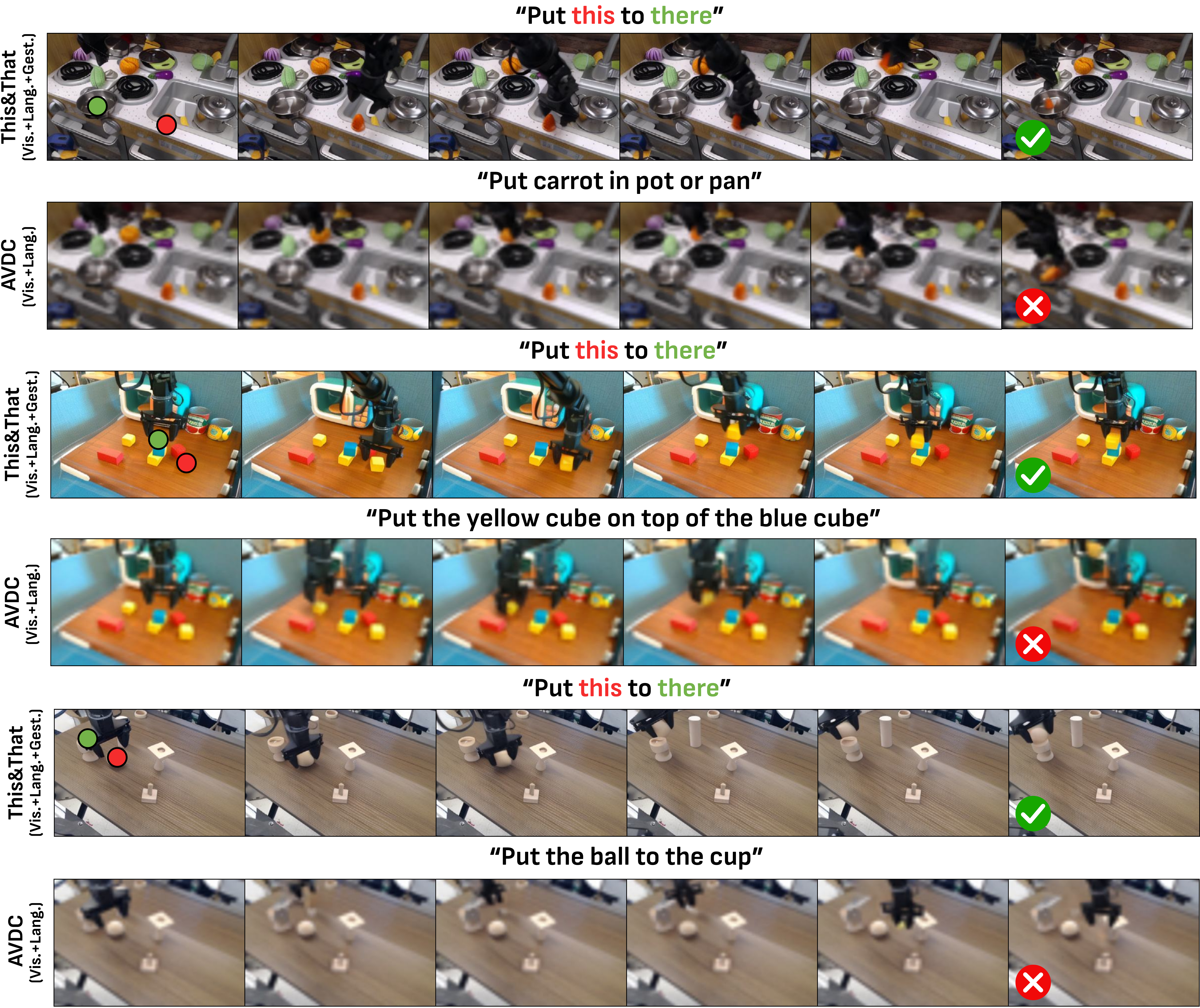} 
    \caption{\textbf{Video-based Planning Qualitative Results.} We present three examples to compare {\em This\&That} with AVDC~\cite{ko2023learning}. The gesture locations are overlayed in the leftmost frame. Our method can generate action sequences effectively with higher visual quality, even when using deictic words.}
    \label{fig:Main_visual}
    \vspace{-4.5mm}
\end{figure*}

\section{Language-Gesture Conditioned Video Diffusion Models}
\label{subsec:videodiffusion}

Video diffusion models (VDMs) learn to remove Gaussian noise from video frames for generative modeling, similar to image diffusion models. Leading VDMs \cite{blattmann2023stable} operate in latent space to reduce computational demands, using encoder $\mathcal{E}$ and decoder $\mathcal{D}$ for conversion between pixel and latent space.

We use pre-trained Stable Video Diffusion (SVD)~\cite{blattmann2023stable} as our foundational latent VDM for open-vocabulary video generation. Our VDM predicts $T$ frames from an initial frame $\textit{I}_0$, sampling from $p_{\theta}(\textit{I}_0,...,\textit{I}_T|\textit{I}_0, C_{\text{text}}, C_{\text{gest}})$ conditioned on language and gesture. We fine-tune SVD first with text and initial frames on a robotics dataset, then modify the architecture and fine-tune it with gesture conditioning.

\subsection{Language-Conditioned Finetuning} 
\label{subsec:vdmfinetuning}

Most large VDMs, such as SVD, are trained on general open-domain datasets and are not directly suitable for robotic tasks. To address this, we retrain SVD’s core structure but initially fine-tune it on robotics videos. We incorporate language description $C_{\text{text}}$ and first frame $\textit{I}_0$ conditionings through Feature-wise Linear Modulation (FiLM)~\cite{perez2018film}, modulating intermediate features using parameters derived from cross-attention between language and image tokens, with token extraction facilitated by the CLIP \cite{radford2021learning} encoder.

During training, we introduce Gaussian noise to frames $\textit{I}_{0:T}$ and optimize the model using the noise reconstruction loss, conditioned on the initial frame and text. Considering the nature of robotics videos, which mark the clear start and end of tasks, we uniformly subsample the sequence to compile $T$ target frames per video.

\subsection{Gesture-Conditioned Training and Inference}
\label{subsec:this_that_cond}

\begin{table*}[t]
  \centering
  \resizebox{0.875\textwidth}{!}{%
    \begin{tabular}{c|r|r|r|r|r|r|r|r|r|r}
      \toprule
      & \multicolumn{2}{c|}{Pick\&Place}
      & \multicolumn{2}{c|}{Stacking}
      & \multicolumn{2}{c|}{Folding}
      & \multicolumn{2}{c|}{Open/Close}
      & \multicolumn{2}{c}{Average} \\
      \cmidrule(lr){2-11}
      Modality
        & Reg & Deic & Reg & Deic & Reg & Deic & Reg & Deic & Reg & Deic \\
      \midrule
      Vision (V.)           
        &  0.0 &   -   &  6.6 &   -   & 11.1 &   -   & 60.0 &   -   & 16.7 &   -    \\
      AVDC (V.+Lang.)       
        &  8.3 & 8.3   &  0.0 & 0.0   &  5.6 & 5.6   & 40.0 & 40.0  & 12.5 & 12.5  \\
      V.+Lang.              
        & 37.5 & 4.2   & 26.7 & 6.6   & 50.0 & 33.3  & \textbf{100.0} & 66.7  & 51.4 & 25.0  \\
      V.+Gesture            
        & 58.3 &   -   & 66.7 &   -   & 55.6 &   -   & \textbf{100.0} &   -   & 68.1 &   -    \\
      V.+Lang.+Gesture      
        & \textbf{95.8} & \underline{91.6}
        & \textbf{80.0} & \underline{66.7}
        & \underline{88.9} & \textbf{94.4}
        & \textbf{100.0} & \underline{93.3}
        & \textbf{91.7} & \underline{87.5} \\
      \bottomrule
    \end{tabular}%
  }
  \caption{\textbf{User Alignment Evaluation of Video Generation.} 
    We conduct a user study to evaluate whether the videos generated from various conditioning modalities align with the ground truth user intent. Success rates are shown in two columns per task (\textbf{Regular Text \textbar{} Deictic Text}). 
    The deictic text indicates the post-processed ambiguous prompt, and regular text means non-deictic text. {\em V.} and {\em Lang.} denote vision and language modalities. For {\em V.} and {\em V.+Gesture}, no text conditioning is used. The best for each task is highlighted and the second best is underlined.}
  \label{table:main_comparison}
  \vspace{-2mm}
\end{table*}

We use a combination of language and pointing gestures, marked as 2D coordinates relative to the first frame, to intuitively control video generation. Each gesture point is converted to image form. To condition video generation on gesture, we adapt our VDM architecture (Fig.~\ref{fig:diffusion}) to include a supplementary network parallel to our fine-tuned diffusion UNet, following ControlNet~\cite{zhang2023adding} in image-conditioned generation. At this stage, we freeze the diffusion UNet weights and train only the new gesture conditioning branch. 

We find that na\"ively applying ControlNet to our case does not ensure that video generation follows the provided gestures. We hypothesize that this is due to the spatial and temporal sparsity of the conditioning information. Specifically, we present two 2D coordinates (1-pixel per coordinate) as gestures for the first two frames, leaving all remaining frames blank without any information. Furthermore, gesture alone does not fully define the task; it needs to be interpreted alongside the text prompt during video generation. 

To resolve these issues, inspired by~\cite{ma2024follow}, in the gesture conditioning branch before executing the first convolution layer, we concatenate the first frame encoded latents $\mathcal{E}(I_{\mathrm{0}})$, noisy video frames $\epsilon_t$ at denoising step $t$, and the sparse gesture images $\mathcal{E}(C_{\text{gest}})$. This simple but effective adjustment ensures dense input signals across the conditioning branch, allowing sparse gesture inputs to be integrated meaningfully with the current video content. We also modulate this gesture conditioning branch with the language prompt condition as in Sec.~\ref{subsec:vdmfinetuning}. Finally, similar to ControlNet, the output of the conditioning branch is added to the decoder of the UNet.

Since the existing robotics video dataset lacks ground truth gesture information, we propose a self-supervised annotation pipeline by detecting and tracking the 2D gripper along with objects of interest, identifying target gesture points. After obtaining these locations, we enhance the spatial signal by applying a 2D Gaussian filter to dilate the points, following~\cite{wang2023motionctrl}. For tasks like opening a door, where only the initial contact point is relevant, we only consider one gesture point.

\section{Video-Conditioned Behavioral Cloning}
\label{subsec:videotorobot}

We want a policy $\pi(\cdot)$ to translate frames $\mathcal{I} = [I_0, \ldots, I_T]$ from a video plan into executable robot actions. For this, we develop DiVA (\textbf{Di}ffusion \textbf{V}ideo to \textbf{A}ction), a BC model based on a Transformer encoder-decoder architecture \cite{vaswani2017attention} that learns to sample from the conditional action distribution $\pi_\theta(a_{t:t+k} | o_t, s_t, \tau)$, where $a_{t+k}$ is the chunk of actions to execute next, $o_t$ is the live image observation, $s_t$ is the robot's end-effector pose, and $\tau$ is a subset of the video plan $\mathcal{I}$.

While DiVA's Transformer-based architecture shares similarities with recent behavior cloning methods like ACT~\cite{zhao2023learning}, DiVA uniquely incorporates predicted future videos for action prediction. Our novel design described below enabled effective video-referencing behavior cloning, a strategy that, to our knowledge, has not been previously explored. 

\begin{figure*}[t]
    \centering
    \includegraphics[width=0.9\textwidth]{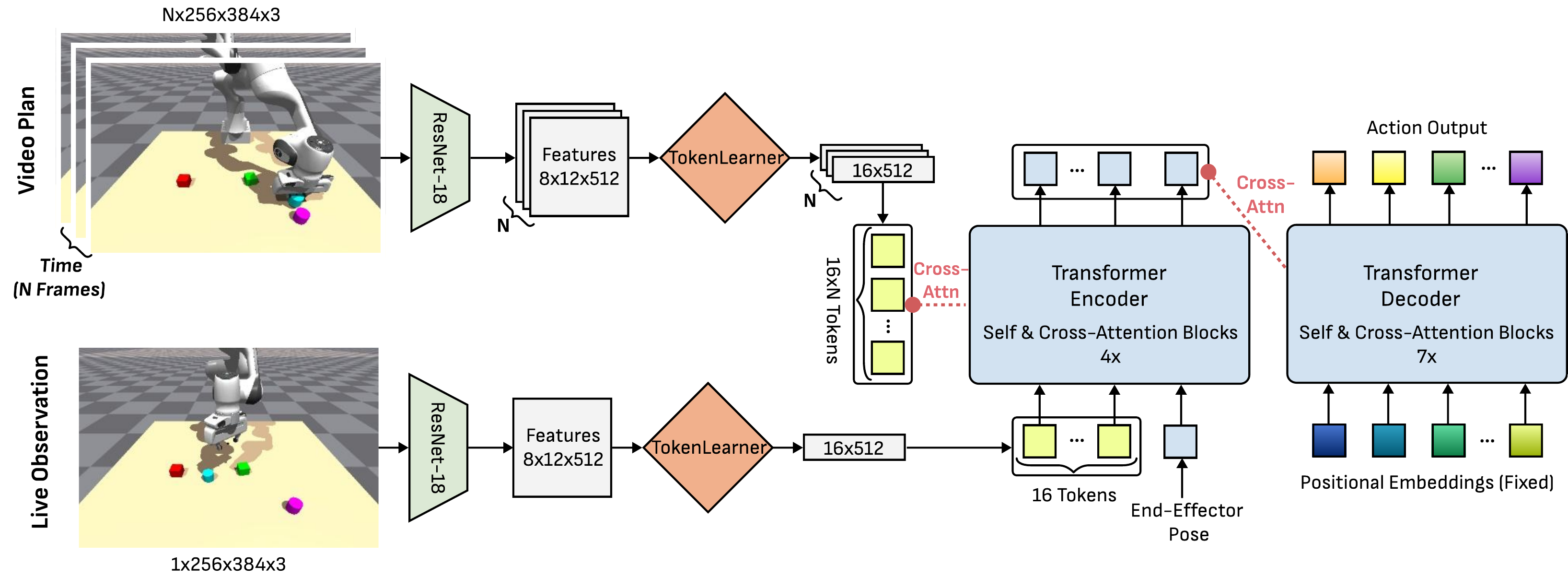}
    \vspace{-2pt}
    \captionof{figure}{\textbf{Video-conditioned Behavior Cloning Architecture.} Our Diffusion Video to Action (DiVA) model utilizes a Transformer encoder-decoder architecture to convert video plans into executable robot actions by compressing image embeddings using TokenLearner~\cite{ryoo2021tokenlearner} and referencing video plan tokens with cross-attention.}
  \label{fig:architecture}
\end{figure*}

Given the video plan subset $\tau$ and live image observation $o_t$, we first convert them into latent embeddings using ResNet-18, pre-trained on ImageNet and fine-tuned during training. Since we are dealing with a potentially large number of images in the video plan, we apply self-attention via TokenLearner \cite{ryoo2021tokenlearner} to compress all image embeddings to just 16 tokens per image. DiVA uses a Transformer encoder to perform cross-attention between live image tokens and tokens from the video plan. Finally, a Transformer decoder is used to convert fixed positional embeddings into a chunk of actions by cross-attending to the encoder output. Fig.~\ref{fig:architecture} provides an overview of our method. 

During training, we use $N$ evenly spaced images from each demo as ground truth goals. In inference, these are replaced by outputs from our VDM. We intentionally choose not to train DiVA on generated frames from our VDM to demonstrate that there is very little domain gap between ground truth and model-generated video. However, because our VDM generates a fixed number of frames and our action sequences vary in length, initial experiments revealed errors likely due to small temporal misalignments. To enhance robustness, we introduce temporal noise during training by randomly sampling an image from $N$ even groups of ground truth observations to form the goal sequence. We ablate $N$ and temporal noise in the supplemental material.

Overall, we attribute DiVA's success to three main factors: 1) compression of latent image embeddings via TokenLearner, 2) cross-attention between observation and goal tokens, and 3) addition of temporal noise when sampling ground truth goals during training. 

\begin{table*}[t]
  \centering
    \small
    \begin{tabular}{l || c | c | c | c | c}
      \toprule
      \textbf{Methods} 
        & FID$\downarrow$ 
        & FVD$\downarrow$ 
        & PSNR$\uparrow$ 
        & SSIM$\uparrow$ 
        & LPIPS$\downarrow$ \\
      \midrule
      SVD~\cite{blattmann2023stable} 
        & 29.49   & 657.49  & 12.47  & 0.334 & 0.391 \\
      StreamingT2V~\cite{henschel2024streamingt2v} 
        & 42.57   & 780.81  & 11.35  & 0.324 & 0.504 \\ 
      DragAnything~\cite{wu2024draganything}  
        & 34.38   & 764.58  & 12.76  & 0.364 & 0.466 \\
      AVDC~\cite{ko2023learning} 
        & 163.93  & 1512.25 & 20.23  & 0.663 & 0.507 \\
      \textbf{Ours} 
        & \textbf{17.28} & \textbf{84.58} & \textbf{21.71} & \textbf{0.787} & \textbf{0.112} \\
      \bottomrule
    \end{tabular}%
  
  \caption{\textbf{Quantitative Video Quality Analysis on the Bridge Dataset.}  
    We report FID (lower is better), FVD (lower is better), PSNR (higher is better), SSIM (higher is better), and LPIPS (lower is better). 
    The best result in each column is highlighted in bold.}
  \label{tab:quantitative_comparison}
  \vspace{-15pt}
\end{table*}

\section{Experiments}
\label{sec:experiment}

We conduct a series of experiments to show the superior user alignment of our {\em This\&That} framework, focusing on the accuracy of video generation and the translation of video plans into robot actions. Specifically, our main experiments aim to 1) show realistic, user-aligned video generation, 2) assess the effectiveness of language-gesture conditioning and 3) evaluate the integration of predicted video plans into downstream policy learning.

\subsection{Video Generation Experiments and Comparisons}
\noindent\textbf{Bridge Dataset.} We evaluate our video generation framework on the Bridge V1~\cite{ebert2021bridge} and V2~\cite{walke2023bridgedata} datasets, which are widely-used real robot datasets with human teleoperated demonstrations. The Bridge datasets feature complex real-world scenes and tasks, focusing on household environments such as kitchens, laundry areas, and desktops. We only use observations from the front view and prune short and long sequences for simplicity of training. Across V1 and V2, we obtain 25,767 videos for initial fine-tuning (Sec.~\ref{subsec:vdmfinetuning}) and 14,735 videos for gesture-conditioned training (Sec.~\ref{subsec:this_that_cond}), filtering out videos where automatic gesture annotation fails.

\noindent\textbf{Evaluating Video Prediction Quality.}
To assess our video prediction quality, we compare against the most recent video synthesis methods in Image-to-Video (SVD~\cite{blattmann2023stable}), Image-Text-to-Video (StreamingT2V~\cite{henschel2024streamingt2v}), and Image-Gesture-to-Video (DragAnything~\cite{wu2024draganything}). We also compare with AVDC~\cite{ko2023learning} (Image-Text-to-Video), the SOTA open-source VDM for robotics trained on the entire Bridge Dataset.


We evaluate these methods on the 646 test videos of the Bridge dataset using the Frechet Inception Distance (FID)~\cite{fid} and Frechet Video Distance (FVD)~\cite{fvd} to measure visual and temporal generation quality. We also compute loss against the ground truth images using pixel-wise and perceptual metrics (PSNR, SSIM, and LPIPS~\cite{zhang2018unreasonable}). As shown in Tab.~\ref{tab:quantitative_comparison}, our language-gesture conditioned VDM demonstrates superior visual realism and temporal alignment against the baselines, including AVDC which has seen these test videos during training.

\noindent\textbf{User Alignment Study.}
We conduct a human study to evaluate the alignment of generated videos with user intent. The participants are randomly selected undergraduate and graduate students in robotics or computer science. Given the initial image, language prompt, and gesture, participants are asked whether the generated video aligns with the user intent. Their answer is a yes or no response.

We compare with AVDC and our custom VDMs with different modality conditions: vision (first frame), language, and gesture. One model weight is concurrently tested in two forms of language: {\em regular} text from the original Bridge dataset and the {\em deictic} format. The test dataset is 24 sequences from Bridge and is categorized into four robotics tasks: {\em pick and place, stacking, folding, {\em and} open or close}. 

The results in Tab.~\ref{table:main_comparison} and Fig.~\ref{fig:Main_visual} demonstrate that our language-gesture conditioned video model outperforms the baselines for all tasks and language variants. Our method shines in tasks with object ambiguity where the language-only baselines fail to produce any reasonable result (Fig.~\ref{fig:Main_visual} rows 3-6), even though the language prompt, to any human, seems sufficient enough to describe the desired task. Furthermore, we show in Tab.~\ref{table:main_comparison} that our method conditioned on deictic langauge is comparable to the baselines conditioned on regular text, indicating that gesture is a powerful form of instruction that works even with simple language commands. For {\em open and close} tasks, all methods work effectively since there is little ambiguity, and language alone suffices.

\begin{figure*}[h]
    \centering
    \includegraphics[width=\textwidth]{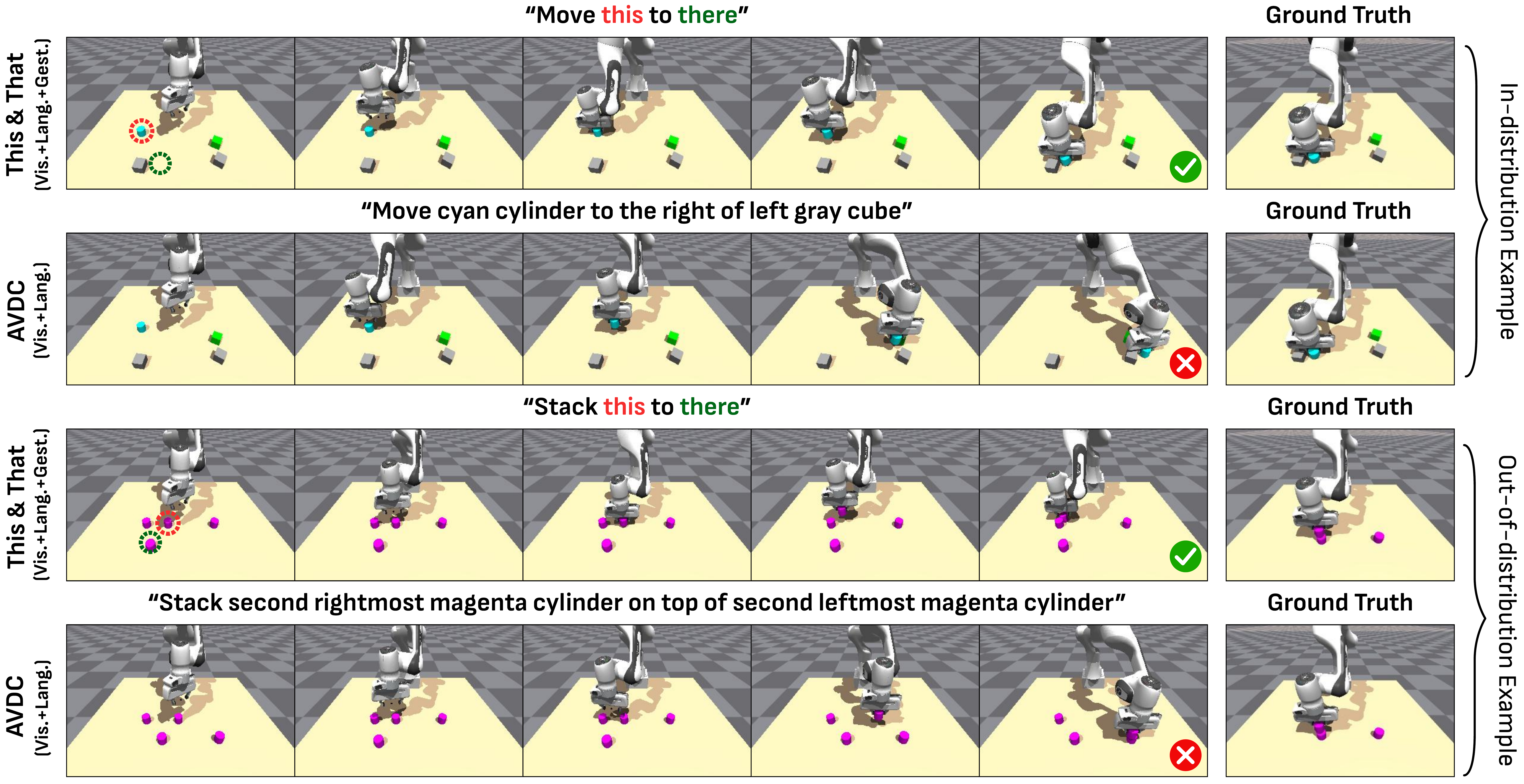}
    \vspace{-2pt}
    \captionof{figure}{\textbf{Simulation Rollout Qualitative Comparison.} We compare the simulated rollouts of our language-gesture conditioned model against AVDC. AVDC struggles to interpret complex text instructions and resolve scene ambiguities. In contrast, our model effectively translates user intent into actions, even with simple language commands.}
    \label{fig:ACT}
    \vspace{-15pt}  
\end{figure*}

\subsection{Synthetic Rollout Experiments}
\label{exp::rollout_result}

\begin{table}[]
\centering
\resizebox{\columnwidth}{!}{%
\begin{tabular}{ccc}
\toprule
 & \multicolumn{2}{c}{Success Rate (\%)}  \\
\cmidrule(lr){1-1} \cmidrule(lr){2-3}
Goal-Conditioning & Pick & Place  \\
\cmidrule(lr){1-1} \cmidrule(lr){2-3}
ACT (V.: Vision-only)               &  5/3 &  0/1 \\
ACT (V.+Lang.)                     &  3/3 &  0/0 \\
ACT (V.+Lang.+Gesture)             & 57/56 & 35/35 \\
AVDC-retrain (V.+Lang.)            & 67/40 & 46/14 \\
Video-based (V.+Lang.)             & 93/60 & 82/26 \\
Video-based (V.+Lang.+Gesture)     & \textbf{95}/\textbf{87} & \textbf{93}/\textbf{80} \\
\bottomrule
\end{tabular}%
}
\caption{\textbf{Quantitative Comparison of Synthetic Robot Rollouts}. We evaluate behavior cloning methods in simulation. The first three rows show ACT \cite{zhao2023learning} variants with different input modalities. The remaining rows compare video generation-based planning baselines, including AVDC \cite{ko2023learning}. ``Pick" success indicates successful grasping and lifting of the target object while ``Place" success indicates pick success and successful movement of the picked object to the target location. Results are presented as \textbf{in-distribution / out-of-distribution} scene performance. Best results are highlighted in bold.}
\label{table:rollout}
\end{table}

We design a simulation environment using Isaac Gym~\cite{makoviychuk2021isaac} to evaluate our method's performance. We set four blocks on a tabletop environment and prepare pick-and-place tasks by relating two randomly selected objects in diverse ways (e.g., {\em stack the blue cube \textbf{on top of} the red cylinder}). We use a hand-scripted policy to obtain ground truth trajectories and record conditionings programmatically. All video models in Tab.~\ref{table:rollout} and Fig.~\ref{fig:ACT} are trained from scratch on visual observations from trajectories collected in simulation. We instantiate the weights of our custom VDMs from SVD. During testing, we stress-test our methods by introducing out-of-distribution scenes that contain identical blocks (same shape and color).

In Tab.~\ref{table:rollout} and Fig.~\ref{fig:ACT}, we compare our full pipeline (language-gesture conditioned VDM + DiVA) against baselines with different combinations of conditioning. First, we show that our method outperforms ACT conditioned on language and gesture (Tab.~\ref{table:rollout} row 3), justifying the use of video generation as an intermediate representation for task planning and execution. Second, we compare our pipeline with AVDC + DiVA (Tab.~\ref{table:rollout} row 4 and Fig.~\ref{fig:ACT}), proving that our method beats existing SOTA language-conditioned video planners for robotics. 
Lastly, we show that our method outperforms our custom VDM conditioned on language (Tab.~\ref{table:rollout} row 5). We conclude that gesture is necessary to generate accurate video plans in uncertain environments.

For fairness of comparison, we modify text prompts at test time by adding absolute spatial adjectives like \textit{front}, \textit{back}, \textit{left}, and \textit{right}, to disambiguate identical blocks. However, we still find language instruction alone is inadequate to complete tasks with ambiguity (Fig.~\ref{fig:ACT} rows 2 and 4). We hypothesize that the pre-trained large language models, such as CLIP or BeRT, attached to our language-conditioned baselines are weak at complex spatial reasoning, which has been widely recognized by the NLP community \cite{cohn2024evaluatingabilitylargelanguage, wang2024pictureworththousandwords}. Our method, in contrast, achieves high success by incorporating gesture. 

\begin{figure*}[t]
  \centering
  \includegraphics[width=\textwidth]{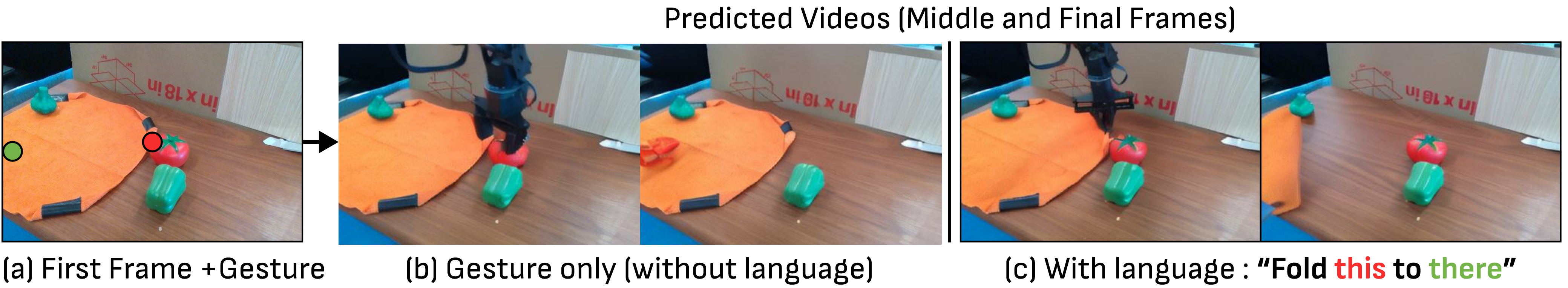}
  \caption{\textbf{Limitation of gesture-only conditioning.} 
    2D gestures can suffer from 3D ambiguity as an image‐plane coordinate does not fully determine a 3D point (see panels (a) and (b)). A simple language cue resolves this ambiguity (panel (c)). Moreover, we empirically observe higher visual quality for language–gesture models, likely due to reduced uncertainty during training.}
  \label{fig:VG}
  \vspace{-1mm}
\end{figure*}

\section{Limitations}
Although our model generates high-fidelity videos, object shapes often change over time, likely due to the lack of 3D geometric constraints. Our predictions are limited to short, modular tasks; extending them to longer tasks (e.g., cooking) with multi-modal instructions presents a significant opportunity. In addition, we highlight a special case in Fig.~\ref{fig:VG} where gesture alone might fail due to 3D ambiguity. \\

\noindent\textbf{Transfer to Real Robots.}
Finally, our video-based behavioral cloning experiments are currently limited to simulated environments. While we achieved state-of-the-art generation results on Bridge videos, the standard real datasets for this task, we could not test with real robots due to the unavailability of the WidowX 250 arm at our academic facility. However, existing research involving the Bridge data has consistently shown the successful transfer of generated video plans to real robot manipulations, and that real-world performance closely correlates with simulated results. Based on these findings, our advancements in video generative models and behavioral cloning represent fundamental contributions to video-based real robot planning.

\section{Conclusion}
In this work, we present {\em This\&That}, a framework that combines the power of visual generative models and imitation learning for effective task communication and planning. 
{\em This\&That} leverages a language-gesture conditioned video generative model as an intermediate planner and uses a video-based behavioral cloning model that elegantly combines the predicted frames and live observation into actions. Our experiments demonstrate that our generated videos and their subsequent rollouts align exceptionally well with user's intentions, suggesting an exciting new direction toward multi-task human-robot collaborations.


\bibliographystyle{IEEEtran}
\bibliography{reference}

\clearpage

\appendix

\subsection{Document Overview}

In this supplemental document, we provide detailed additional content that complements the main paper. Sec.~\ref{sec:addition_exp_ablation} elaborates on additional qualitative results and ablation studies for both our proposed \textbf{V}ideo \textbf{D}iffusion \textbf{M}odel (VDM) and \textbf{Di}ffusion \textbf{V}ideo to \textbf{A}ction model (DiVA); Sec.~\ref{sec:videodiff} details our proposed VDM architecture, training setup, and automatic gesture labeling methods; Sec.~\ref{sec:diva_details} describes the implementation and training specifics of DiVA; Sec.~\ref{sec:experiment_details} covers the experiments performed for both VDM and DiVA—including our human study and simulated rollouts; finally, Sec.~\ref{sec:limitation_VDM} presents the limitations of our VDM.

\subsection{Additional Experiments and Ablation Studies}\label{sec:addition_exp_ablation}

\subsubsection{Qualitative Comparison with Contemporary Video Generative Models}
\begin{figure*}[t]
    \centering
    \includegraphics[width=0.85\textwidth]{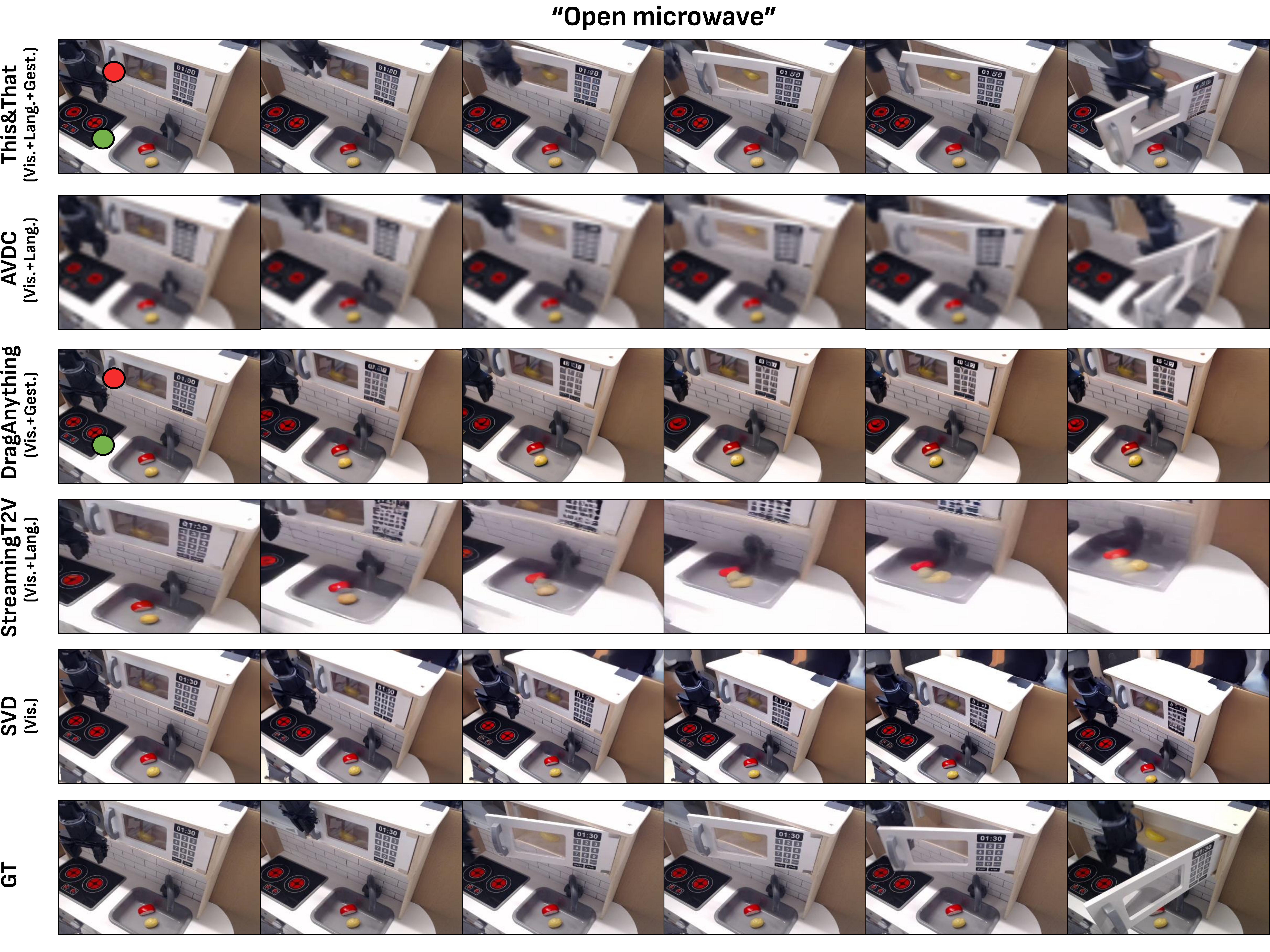} 
    \caption{\textbf{Qualitative Comparison with State-of-the-art Video Generative Models.} Recent methods such as DragAnything~\cite{wu2024draganything}, StreamingT2V~\cite{henschel2024streamingt2v}, and SVD~\cite{blattmann2023stable} fail to generate correct videos following the language or gesture commands. AVDC~\cite{ko2023learning} produces a reasonable result (this scene is part of its training data) but lacks visual quality. {\em This\&That} produces a high-quality video that adheres to the user's intention. }
    \label{fig:video_gen}
    \vspace{-1em}
\end{figure*}

In Fig.~\ref{fig:video_gen}, we present a visual comparison of our method against contemporary video generation models, augmenting the quantitative data presented in Tab.~\ref{table:main_comparison}.

AVDC~\cite{ko2023learning} is trained on the entire Bridge dataset, so it produces a semantically correct sequence. However, the visual quality of AVDC's output is lacking, characterized by low spatial and temporal resolution along with visual artifacts on the microwave door. These deficiencies hinder the accurate interpretation of the end-effector and environmental states, which are critical for translating the videos into robot actions. Other leading video generation models, such as DragAnything~\cite{wu2024draganything}, StreamingT2V~\cite{henschel2024streamingt2v}, and SVD~\cite{blattmann2023stable}, when used directly without specific fine-tuning, were unable to adhere to the provided text or gesture commands. This underscores the need for a specialized language-gesture VDM, specifically designed for robotic applications.

\begin{figure*}[t]
  \centering
  \includegraphics[width=0.85\textwidth]{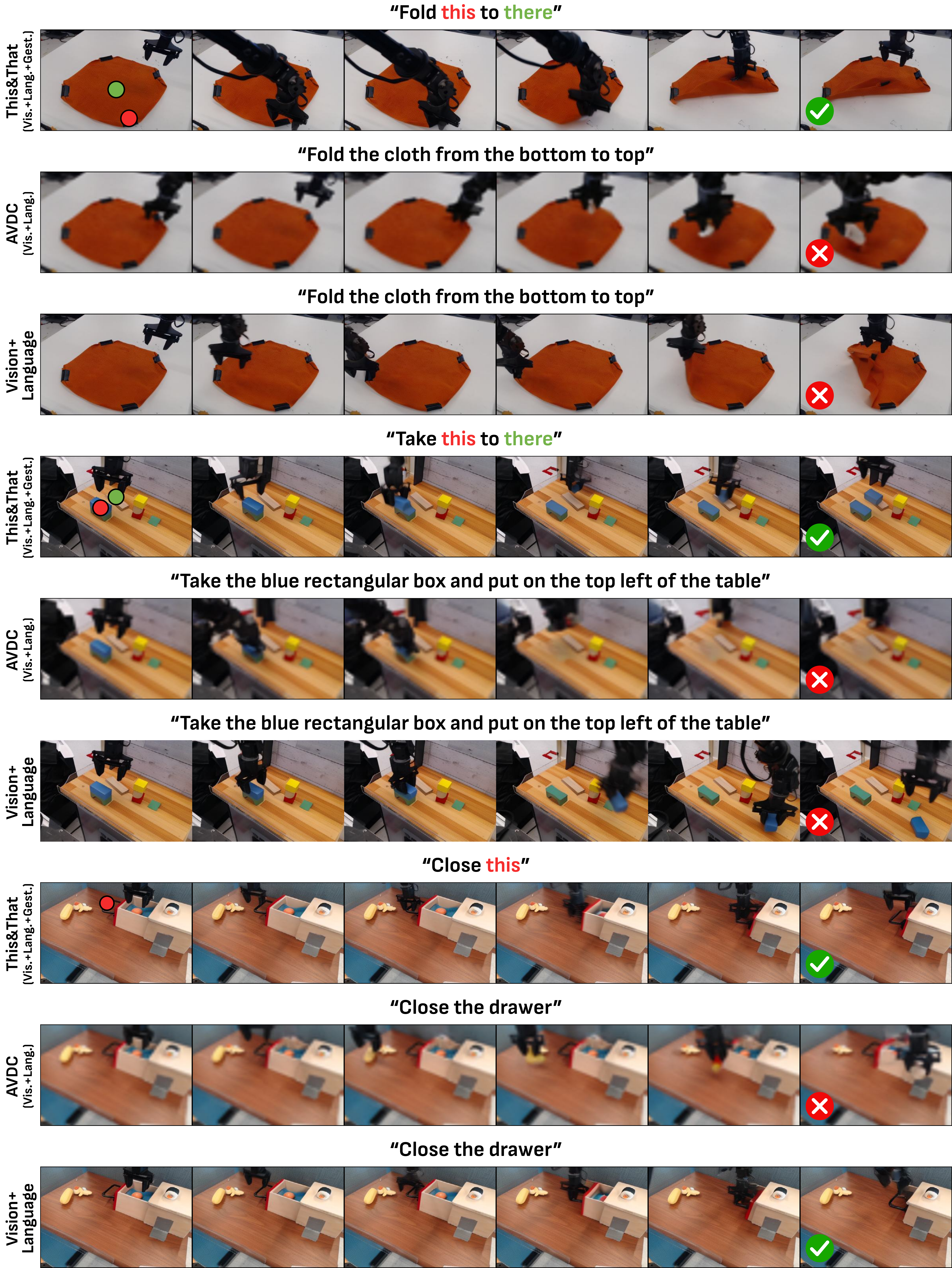}
  \caption{\textbf{Video-based Planning Qualitative Results.} Three examples compare This\&That with AVDC~\cite{ko2023learning} and our VDM baseline conditioned on the first frame and language. Gesture locations are overlaid in the leftmost frame. Note the superior quality and alignment of our generated videos compared to the baselines.}
  \label{fig:vdm_qual}
\end{figure*}

In Fig.~\ref{fig:vdm_qual}, we present additional qualitative results from our test split of the Bridge dataset. We use the provided initial frame and construct new text and gesture prompts to generate unique videos. 
It's important to note that the combinations of prompts and frames in Fig.~\ref{fig:Main_visual} and the first example in  Fig.~\ref{fig:vdm_qual} do not exist in the original dataset, and there are no corresponding ground truth videos. 
This approach was chosen because the entire Bridge dataset, including our test split, was used for training AVDC, which is the current state-of-the-art open-source video generative model for robotics application. 
Our results are compared against both our language-only baseline and AVDC.

In the first example from Fig.~\ref{fig:vdm_qual}, video generative models relying solely on language instruction (3rd row) struggle to capture the nuanced geometric relationship implied by the prompts, resulting in implausible video outputs. The second example reveals that both our language-only baseline (6th row) and AVDC (5th row) were deemed unsuccessful in a user study. Although AVDC attempted to align with the specified direction on the table, the generation quality was poor, and the blue box became invisible after movement. In the third example, our language-only baselines (9th row) performed well, correctly capturing the straightforward ``closing" action. However, AVDC (8th row) failed to follow the text prompt accurately, likely due to overfitting on the training data.

\subsubsection{VDM Ablation Study}
To validate the design decisions behind our Video Diffusion Model (VDM) architecture, we conducted an ablation study from four distinct perspectives (as illustrated in Tab.~\ref{tab:vdm_ablation}).

\paragraph{Assessing Na\"ive ControlNet Conditioning}

Our first ablation revisits the standard ControlNet conditioning architecture. We substituted our proposed approach, which involved pre-trained VAE encoding, with a simple zero convolution, and removed our concatenation method for integrating gesture conditioning and VDM noise inputs. This change significantly reduced performance across all assessed metrics. Visual inspections of the generated videos confirmed that they failed to accurately follow the specified gesture cues, underscoring the superiority of our original concatenation and encoding methods in maintaining adherence to gesture inputs.

\begin{table*}[t]
	\centering
    \small
		\begin{tabular}{l||c|c|c|c|c}
			\hline
			\textbf{Methods} & FID$\downarrow$ & FVD$\downarrow$ & PSNR$\uparrow$ &SSIM$\uparrow$ &LPIPS$\downarrow$ \\
			\hline\hline
	Regular Controlnet                       & 22.158 & 124.710 & 19.975 & 0.758 & 0.134 \\	
    With SAM Segmentation Mask               & 17.922 & 88.757  & 21.554 & 0.785 & 0.115 \\ 
    No Layernorm on CLIP Embeddings     & 17.566 & 92.527  & 21.559 & 0.786 & 0.114 
    \\        
    Larger Gesture Conditioning & 18.844 & 96.794 & 21.180 & 0.778 & 0.122
    \\
    Smaller Gesture Conditioning & 19.813 & 106.953 & 21.506 & 0.782 & 0.119
    \\
    \textbf{Ours}           & \textbf{17.278} & \textbf{84.580} & \textbf{21.716} & \textbf{0.787} &\textbf{0.112} 
    \\
    
			\hline
	   \end{tabular}
        
        \vspace{1.5mm}
        \caption{\textbf{Ablation study on VDM}. The precision is 3 digits after the decimal point. The best is highlighted.}
	\label{tab:vdm_ablation}
\end{table*}

\begin{figure*}[t]
    \centering
    \includegraphics[width=0.8\textwidth]{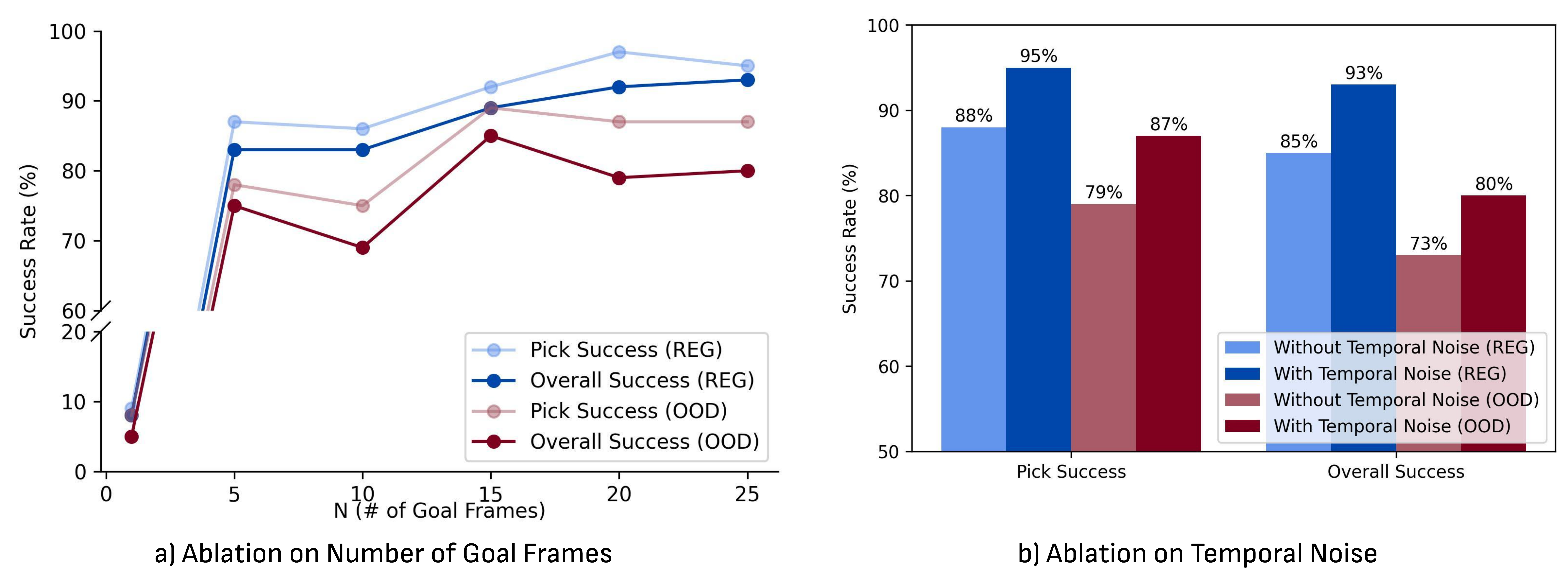} 
    \caption{\textbf{DiVA Ablation Studies.} We ablate DiVA on a) the number of goal frames (N) and b) the addition of temporal noise during training. DiVA overall performs best with 25 goal frames and temporal noise.}
    \label{fig:rollout_ablation}
\end{figure*}

\paragraph{Using Semantic Segmentation Masks in Gesture-Conditioning}
The second ablation experiment investigates the usage of segmentation masks during training, which could potentially offer more spatial information during gesture-conditioned training. By querying the SAM~\cite{kirillov2023segment} with the ``pick" gesture location, we can obtain a segmentation mask that provides denser gesture signals. However, this additional spatial information does not translate to improved numerical performance. The segmentation algorithm often misinterprets the intended objects, outputting broader scene segments rather than specific objects, like outputting a desk instead of just the cup on it. This inclusion of extraneous pixel data complicates effective training. Consequently, we found that a simpler approach using 2D dilation from a single point yields better results than employing SAM masks.

\paragraph{Language and First Frame Conditioning}
Next, we focus on the analysis concerning the cooperation between language and image embeddings using FiLM and cross-attention. Specifically, we empirically find out that the layer normalization layer after concatenating the CLIP embeddings plays an important role in maintaining the video quality: i.e., removing the layer resulted in poorer performance across all numerical metrics. 

\paragraph{Influence of the Area of Gesture Conditioning}
To enhance the gesture conditioning from one-pixel 2D coordinates to an area with exaggerated color marks, we apply the outpainting method to create a 10x10 pixel block. However, a concern is whether the area of this hand-crafted region will influence the performance. Hereby, we make another ablation study to analyze its influence. In this ablation study, we consider two directions of influence: increasing (double) and decreasing (halve) the area. As shown in Tab.~\ref{tab:vdm_ablation}, we find that applying a gesture area that is too large or too small is not very helpful.

\subsubsection{Additional Quantitative Comparison Considering Image Resolution}
\label{sec:exp_image_resolution}
In Tab.~\ref{table:main_comparison}, all metrics are compared after being resized to 256x384, the default resolution for our VDM. However, AVDC outputs frames at a much lower resolution of 48x64. Since AVDC is very memory intensive and we are limited to a 48GB Nvidia L40S device, we cannot re-train their VDM to the same resolution as ours. Instead, for fairness of comparison, we choose to resize GT images and frames from our VDM to match AVDC's resolution. As shown in Fig.~\ref{tab:vdm_extra}, these additional comparisons still assert that our method outperforms AVDC in all metrics.

\subsubsection{Additional Qualitative Results from Simulation Rollouts}

In Figures~\ref{fig:rollout_supp_2} and~\ref{fig:rollout_supp_1}, we present further qualitative results from our simulated experiments in Isaac Gym~\cite{makoviychuk2021isaac}. Specifically in Fig.~\ref{fig:rollout_supp_2}, we compare the video plans generated by our method, AVDC, and the video-based baseline conditioned solely on language. Then, in Fig.~\ref{fig:rollout_supp_1}, we show the rollouts after using DiVA to execute the video plans. 

These results showcase the ability of {\em This\&That} to generate video plans and effectively translate them into robotic actions. While language-conditioned (without gesture) VDM baselines generally perform well, they struggle in uncertain scenes with identical blocks. To handle this ambiguity, we augment the language prompt at test time by adding adjectives to specify absolute location. For instance, the language prompt for the first example in Fig.~\ref{fig:rollout_supp_2} is modified from ``stack green cube on top of green cube'' to ``stack \textit{right} green cube on top of \textit{left} green cube'' because there are two identical green cubes in the scene. But even with more detailed language, the baselines still fail to resolve scene ambiguity. They perform especially poorly on our out-of-distribution scenes that contain all identical blocks, as reported in Tab.~\ref{table:rollout}. {\em This\&That}, meanwhile, leverages gesture to generate high-quality videos that align with user intent.

\subsubsection{Ablating DiVA: Number of Goal Images and Temporal Noise}\label{sec:diva_ablation}

We perform two ablation studies for DiVA as shown in Fig.~\ref{fig:rollout_ablation}. For the first study, we vary N, the number of goal frames from 1 to 25 in increments of 5. We see almost no success for N=1 (conditioning on just the last frame). The performance increases roughly linearly with the number of conditioned frames, but it plateaus at around N=15 until N=25. We hypothesize that more frames will be beneficial for complex tasks and leave conducting such challenging experiments as future work. 

For the second study, we analyze the effects of adding temporal noise during training. In other words, we split the GT observation images for a demo into N consecutive groups and randomly sampled one image from each group to use in the goal sequence. We find that implementing temporal noise makes DiVA more accurate and robust. We compare the pick and overall success rates of DiVA with and without temporal noise in Fig.~\ref{fig:rollout_ablation}.

\subsection{Video Diffusion Model Implementation Details}\label{sec:videodiff}

\subsubsection{Base Architecture}

The video diffusion model (VDM) we use is based on the Stable Video Diffusion (SVD) framework~\cite{blattmann2023stable}, which incorporates a modified version of the denoising algorithm from the EDM~\cite{karras2022elucidating}, a continuous-time diffusion model framework.
Since the training code of SVD is not publicly available, we first deploy an open-source codebase~\footnote{\href{https://github.com/pixeli99/SVD_Xtend}{https://github.com/pixeli99/SVD\_Xtend}} and make several modifications to the denoise algorithm as described in the following sections.

\subsubsection{UNet Finetuning Details (Stage 1)}
The SVD framework governs video motion using two key parameters: the motion bucket ID and noise augmentation. In robotics applications, a complete video sequence is crucial, depicting the robot arm completing its task. Consequently, we set the motion bucket ID to 200 and the noise augmentation to 0.1, both during training and inference, to override motion control from the pretrained model.

Building on the method proposed by ~\cite{dai2023animateanything}, we enhance the stability of our VDM by discarding a small amount of noise $\log \sigma \sim N(-3.0, 0.5^2)$ traditionally added to the conditioning frame. Instead, we introduce a fixed noise value of 0.1 as an augmentation during training.

For effective text and first frame image conditioning, we concatenate the embeddings prior to their introduction to the encoder hidden states. Given the varying dimensions of text embeddings across different open-source CLIP~\cite{radford2021learning} models, we select a version that matches the feature dimension of our SVD's CLIP image embeddings, which is 1024. We utilize a CLIP encoder from the StableDiffusion2.1 framework ~\cite{rombach2022high} for text embeddings, resulting in dimensions of $x_{\mathrm{text}} \in \mathbb{R}^{b \times 77 \times 1024}$ and $x_{\mathrm{I_0}} \in \mathbb{R}^{b \times 1 \times 1024}$ for image embeddings, aligning perfectly for concatenation. The final concatenated dimension is $x_{\mathrm{concat}} \in \mathbb{R}^{b \times 78 \times 1024}$. We observe that applying Layer Normalization \cite{ba2016layer} to these concatenated embeddings enhances both the visual and quantitative outcomes of our model as shown in Tab.~\ref{tab:vdm_ablation}. The processed language and vision encoder hidden states for the diffusion model are then defined as:
    $\mathrm{{y}_{hs}} = \mathrm{LayerNorm}(
    [\text{CLIP}(x_{\mathrm{text}}) 
    ;
    \text{CLIP}(x_{\mathrm{I_0}})]
    ).$

\begin{figure*}[t]
    \centering
    \includegraphics[width=1.0\textwidth]{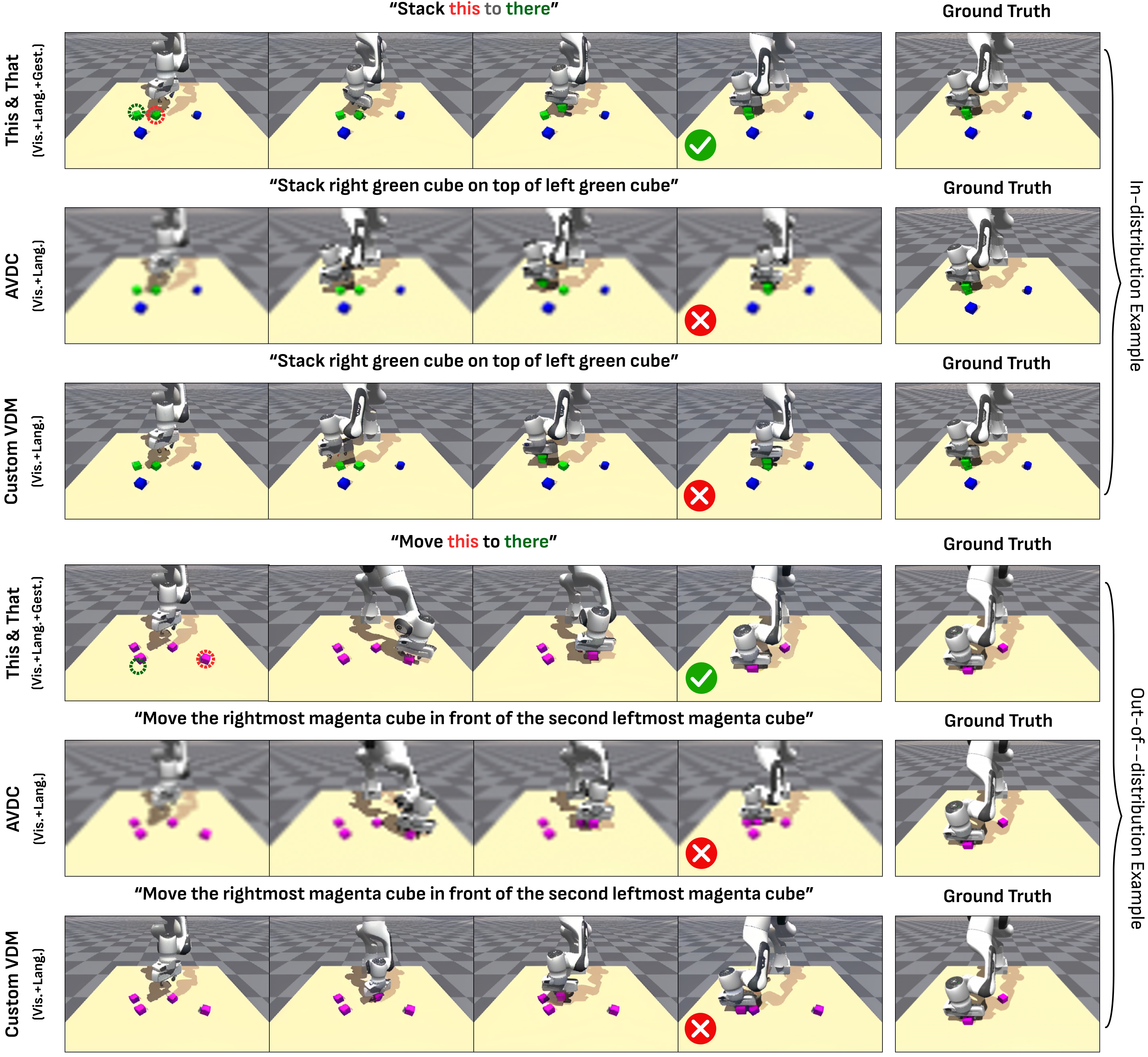} 
    \caption{\textbf{Simulation Video Plan Qualitative Comparison.} We compare video plans from our language-gesture model against AVDC and the video-based baseline conditioned solely on language. The language-only baselines struggle to interpret complex text prompts and resolve scene ambiguities. In contrast, our model can effectively follow user intent.}
    \label{fig:rollout_supp_2}
\end{figure*}

\begin{figure*}[t]
    \centering
    \includegraphics[width=1.0\textwidth]{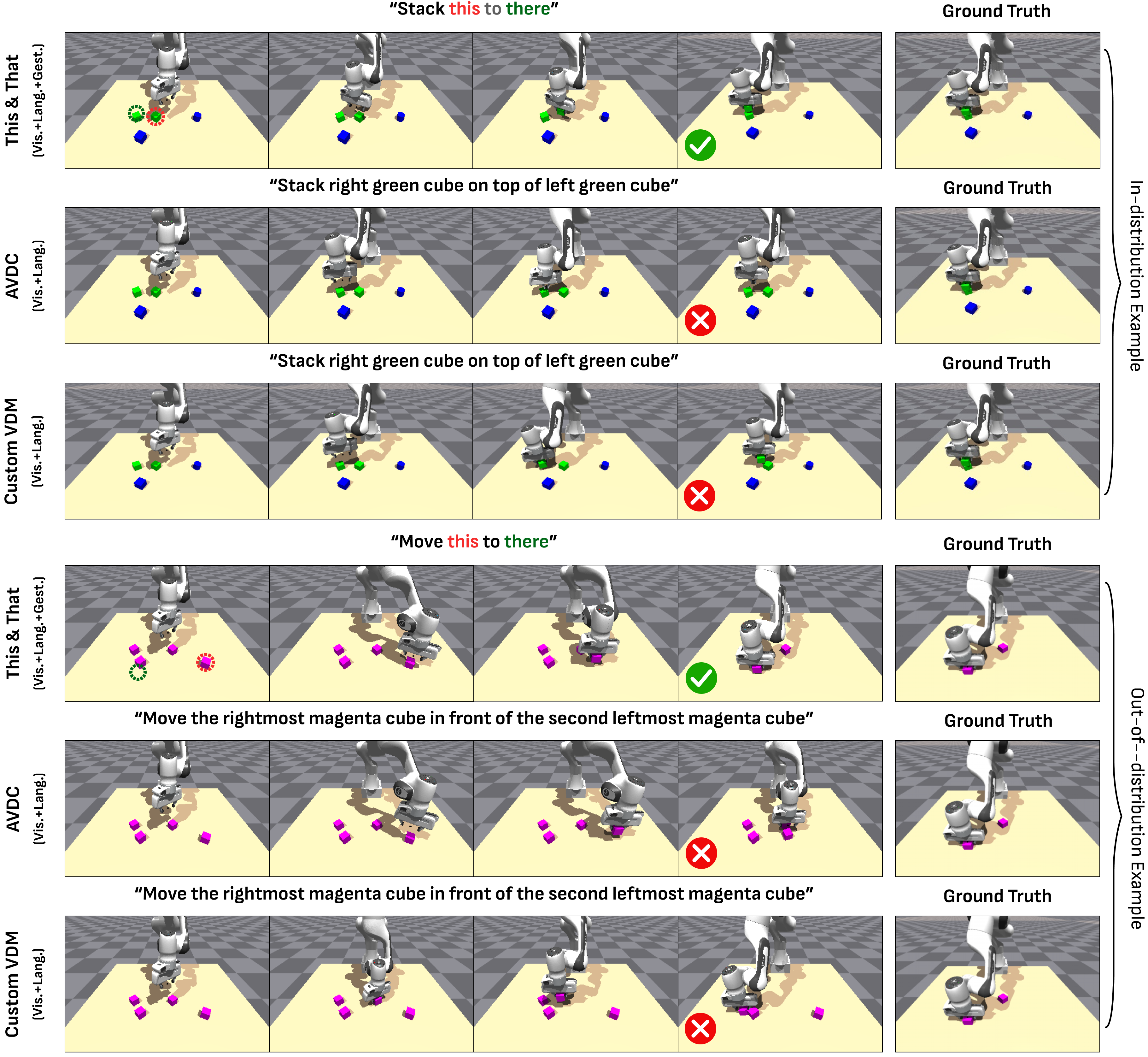} 
    \caption{\textbf{Simulation Rollout Qualitative Comparison.} We compare simulated rollouts, generated by taking actions from DiVA, of our language-gesture model against AVDC and the video-based baseline conditioned solely on language. Rollouts from our method are more accurate than any of the language-only baselines.}
    \label{fig:rollout_supp_1}
\end{figure*}

\begin{figure*}[t!]
\centering
    \includegraphics[width=0.75\textwidth]{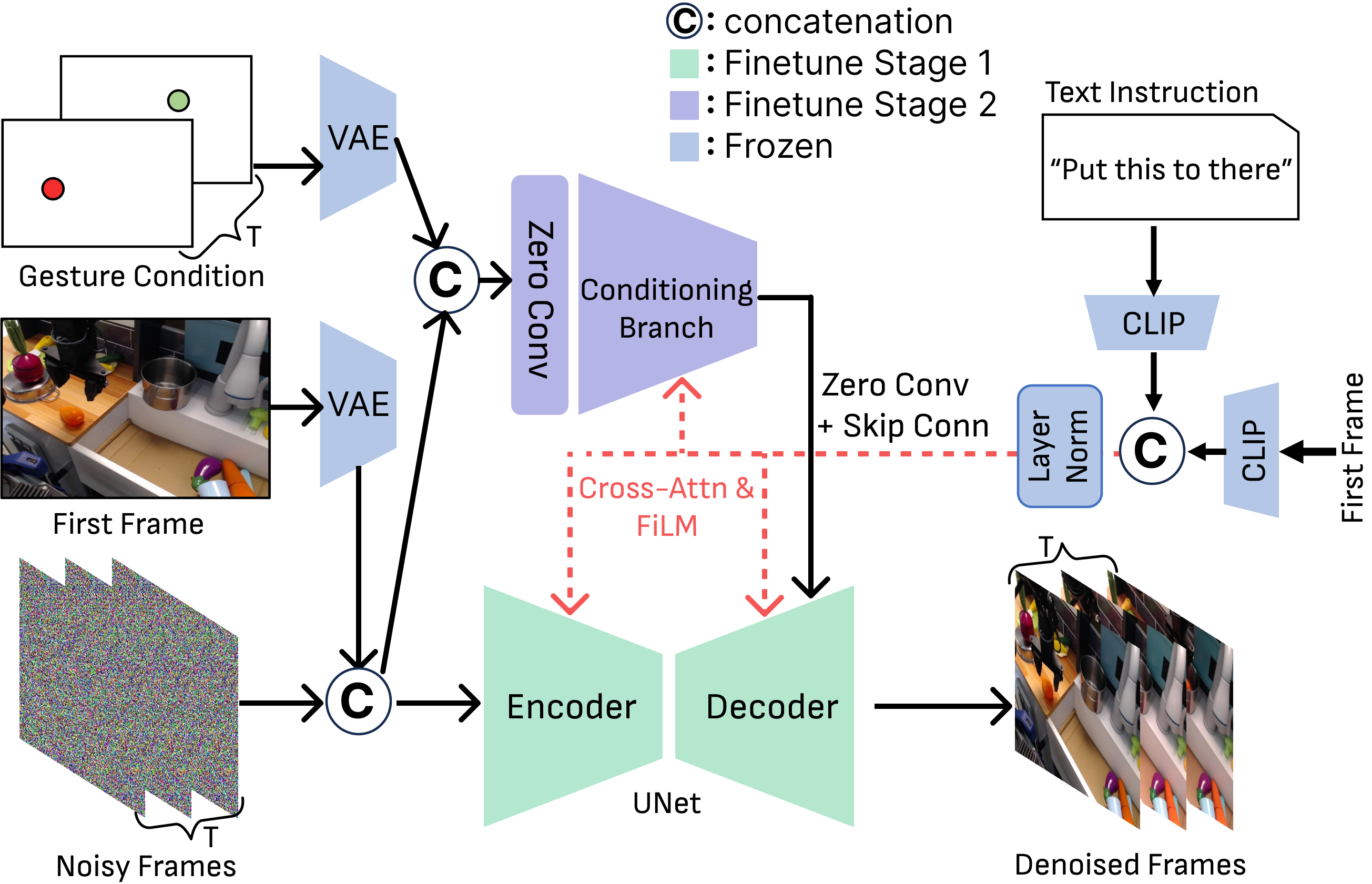}
    \captionof{figure}{\textbf{A detailed visualization of our VDM architecture.} In the first fine-tuning stage, we fine-tune the UNet weights of SVD \cite{blattmann2023stable}, conditioned on the first frame and the CLIP \cite{radford2021learning} embedding of the text inputs. In the second stage, we initialize the conditioning branch using the weights from the UNet encoder and attach a zero convolution layer whose weights are initialized to be zero, following ControlNet \cite{zhang2023adding}. Note that the gesture conditioning images are mapped into the latent space and concatenated with the noise and first frame latents. The CLIP embeddings of the first frame and the text prompts are concatenated and processed with layer normalization before being used to modulate the UNet and the conditioning branch with cross-attention and FiLM \cite{perez2018film} operation.}
  \label{fig:specific_architecture}
\end{figure*}

\subsubsection{Gesture Conditioning Branch Details (Stage 2)}

In the second stage of training for the gesture conditioning branch, we implement temporal conditioning by distributing two gesture points across different frames. For a sequence of $T$ target frames, only two frames contain gesture point pixel information, while the remaining frames hold zero-value images as placeholders. The frames with gesture points are positioned randomly within the sequence, ensuring that the first (red) point appears temporally before the second (green) point as shown in Fig.~\ref{fig:specific_architecture}. Each gesture point is represented as a square box, 10 pixels per side, centered on the specified coordinate. Building on the techniques from MotionCTRL~\cite{wang2023motionctrl}, we employ a 2D Gaussian dilation method to enhance the visibility of gesture points. 

The inputs for the first convolution layer include noise vector $\mathcal{\epsilon} \in \mathbb{R}^{(B\times T) \times 4 \times H \times W}$, the encoded initial frame repeated $T$ times $\mathcal{E}(C_{\mathrm{0}}) \in \mathbb{R}^{(B\times T) \times 4 \times H \times W}$, and the gesture condition vector $\mathcal{E}(C_{\mathrm{gest}}) \in \mathbb{R}^{(B\times T) \times 4 \times H \times W}$. The dimensions $(B\times T)$, $H$, and $W$ indicate batch size times number of frames, and the height and width of the latent space, respectively. The resulting channel-wise concatenation forms a shape of $\mathbb{R}^{(B\times T) \times 12 \times H \times W}$.

Given the discrepancy in input dimensions between the gesture conditioning branch and the UNet's first convolution layer, we do not reuse the pre-trained convolution layer from the UNet. Instead, we train a new convolution layer from scratch with zero initialization (zero convolution), which helps mitigate potential issues from random noise gradients during the initial stages of training. Following the concept from Follow Your Clicks ~\cite{ma2024follow}, we implement latent masking on the gesture conditioning inputs by turning latent features off with some probability $p$. This masking technique is applied exclusively on the gesture conditioning side to prevent the performance degradation observed when applied on the UNet side. 

The input formulations for the gesture conditioning and UNet in stage two are as follows:
\[
\mathrm{{y}_{Gest}} = 
\mathcal{Z}([
\mathcal{M}_p(\mathcal{E}(I_{\mathrm{0}}))
;
\mathcal{\epsilon} 
;
\mathcal{E}(C_{\mathrm{gest}})
]),
\]
\[
\mathrm{{y}_{UNet}} = 
\mathcal{C}([
\mathcal{E}(I_{\mathrm{0}})
;
\mathcal{\epsilon} 
]),
\]
where $\mathcal{Z}(\cdot)$ represents the zero convolution, $\mathcal{C}(\cdot)$ represents the regular convolution layer, and $\mathcal{M}_p(\cdot)$ represents the latent mask with probability $p$. The $\mathrm{{y}_{Gest}}$ vector serves as the input to the conditioning branch encoder, while $\mathrm{{y}_{UNet}}$ feeds into the UNet encoder. Note that both the UNet and the gesture conditioning branch intermediate features are modulated using FiLM~\cite{perez2018film} conditioning, whose parameters are obtained by cross-attending to the CLIP embeddings. We refer readers to the SVD and StableDiffusion implementations for details of this structure. Finally, the conditioning branch is used to provide skip connection values to the UNet decoder following the ControlNet architecture \cite{zhang2023adding}.

\subsubsection{Automatic Gesture Labeling on Real Data}
The Bridge datasets~\cite{ebert2021bridge, walke2023bridgedata} provides metadata of the robot end-effector actions, from which we can recover the key moments when the robot end-effector either closes to grasp or reopens to release objects. Utilizing these temporal markers, our objective is to precisely determine the gripper's interaction points with objects at these key frames. To facilitate this, we employ a bounding box detector developed through training a YoloV8~\cite{redmon2016you} model on 450 manually annotated images that mark the gripper's location. This specialized training enables the model to accurately outline the gripper's bounding box in subsequent frames, ensuring automatic detection and tracking of its interactions with objects.

With the gripper's bounding box identified, we extract the 2D coordinates of the objects at the target frame index. We then employ the TrackAnything model~\cite{yang2023track} to track the objects' motion over time. This tracking approach is crucial, especially in scenarios where objects are released mid-air by the gripper, necessitating continuous monitoring of their trajectory to determine their landing points. This method reliably provides the two necessary gesture points for our application. Further, to decrease the spatial sparsity in the gesture conditioning image format input, we outpaint the single 2D coordinates to a 10x10 pixel block with a mono-color, like green or red.

However, this automated annotation method is not infallible. Challenges arise in tracking more complex interactions, and the Yolo model does not achieve perfect detection accuracy. Consequently, we exclude results from cases where tracking accuracy falls below acceptable thresholds. Additionally, we discard videos that exceed five times the length of the target frame count $T$ or have fewer frames than $T$.

To aid the research community in enhancing gripper detection capabilities, we will release both the detection code and the pre-trained weights. For further details and to access these resources, we encourage interested readers to consult our code repository once it is released. 

\begin{table*}[t]
  \centering
  \small
  \begin{tabular}{@{}lcccccc@{}}
    \toprule
    \textbf{Methods}             & \textbf{FID}↓    & \textbf{FVD}↓    & \textbf{PSNR}↑   & \textbf{SSIM}↑  & \textbf{LPIPS}↓ \\ 
    \midrule
    AVDC – Resize to 256×384     & 163.93           & 1512.25          & 19.43            & 0.649           & 0.517           \\
    Ours – Default 256×384       & \textbf{17.28}   & \textbf{84.58}   & \textbf{21.72}   & \textbf{0.787}  & \textbf{0.112}  \\
    AVDC – Default 48×64         & 62.95            & 620.46           & 22.33            & 0.836           & 0.242           \\
    Ours – Resize to 48×64       & \textbf{24.78}   & \textbf{129.74}  & \textbf{23.13}   & \textbf{0.862}  & \textbf{0.154}  \\
    \bottomrule
  \end{tabular}
  \vspace{1mm}
  \caption{\textbf{VDM Comparisons Considering Image Resolution.} Our method outperforms AVDC across all metrics, even when compared at AVDC’s native low resolution.}
  \label{tab:vdm_extra}
\end{table*}

\subsection{DiVA Model Implementation Details}\label{sec:diva_details}

Our Diffusion Video to Action (DiVA) framework is designed to model the distribution $\pi_\theta(a_{t:t+k} | o_t, s_t, \tau)$, where $a_{t:t+k}$ represents the action chunk the robot executes from time $t$ to $t+k-1$. Here, $o_t$ is the image observation of the environment at time $t$, $s_t$ is the pose of the robot's end-effector at time $t$, and $\tau$ includes a subset of goal images from the video diffusion model output $\mathcal{I}:=I_{0:T}$. We denote $N=|\tau|$ to be the size of the subset. DiVA processes each image of resolution $\mathbb{R}^{256\times384\times3}$ through a ResNet-18 architecture, pre-trained on ImageNet and adjusted during training, to produce latent embeddings of shape $\mathbb{R}^{8\times12\times512}$. 

The latent embeddings for $o_t$ and $\tau$ are then flattened to $z_{o_t} \in \mathbb{R}^{96\times512}$ and $z_{\tau} \in \mathbb{R}^{N\times96\times512}$, respectively. To incorporate the temporal dimension of $\tau$, fixed 2D sinusoidal positional encodings of dimension $\mathbb{R}^{N\times512}$ (expanded to $\mathbb{R}^{N\times96\times512}$) are added to $z_{\tau}$. The embeddings $z_{o_t}$ and $z_{\tau}$ are subsequently processed by a TokenLeaner \cite{ryoo2021tokenlearner} module, which utilizes spatial attention to select 16 dynamic and significant tokens per feature map, resulting in goal tokens $G \in \mathbb{R}^{N\times16\times512}$ and 16 tokens of dimension 512 for $o_t$. TokenLearner is implemented as a multilayer perceptron (MLP) with LayerNorm and a single hidden layer of 64 units. 

Another MLP is used to encode the end-effector pose $s_t$ into a single token. This token is concatenated with $o_t$ to form $O \in \mathbb{R}^{17\times512}$. DiVA then employs a transformer encoder to process $O$ input with 4 layers of self-attention and cross-attention to the goal tokens $G$. This is followed by a transformer decoder applying self-attention to fixed 2D sinusoidal positional embeddings of shape $\mathbb{R}^{k\times512}$ across 7 layers of self and cross-attention to the encoder outputs.

The final $k$ output tokens of the transformer are decoded into the actionable sequence $a_{t:t+k}$ using a standalone MLP for each token. The full action chunk of size $k$ is executed before querying the model again. In our experiment, we use $k=10$. We utilize an L1 loss function for training, which ensures greater stability compared to L2 loss. A schematic overview of DiVA’s architecture is detailed in Fig.~\ref{fig:architecture}.

\subsubsection{Alternative Method: Inverse Dynamics Model}

Initially, we explored training an inverse dynamics model to interpolate actions between frames produced by our video diffusion model, inspired by the methods described in UniPi \cite{du2024learning}. However, we observe that such a strategy failed to produce reasonable action outputs, likely because our diffusion model generates a fixed number of frames for each demonstration, whereas the actual demonstrations vary in length. This discrepancy made it impractical to train an inverse dynamics model that could uniformly output a fixed number of actions for each frame sequence without introducing an additional diffusion model dedicated to temporal interpolation as used in UniPi\cite{du2024learning}. Our method avoids such additional generative steps and is robust to the number of input images and temporal misalignments as shown in Sec.~\ref{sec:diva_ablation}.

\begin{figure*}[t]
    \centering
    \includegraphics[width=0.95\textwidth]{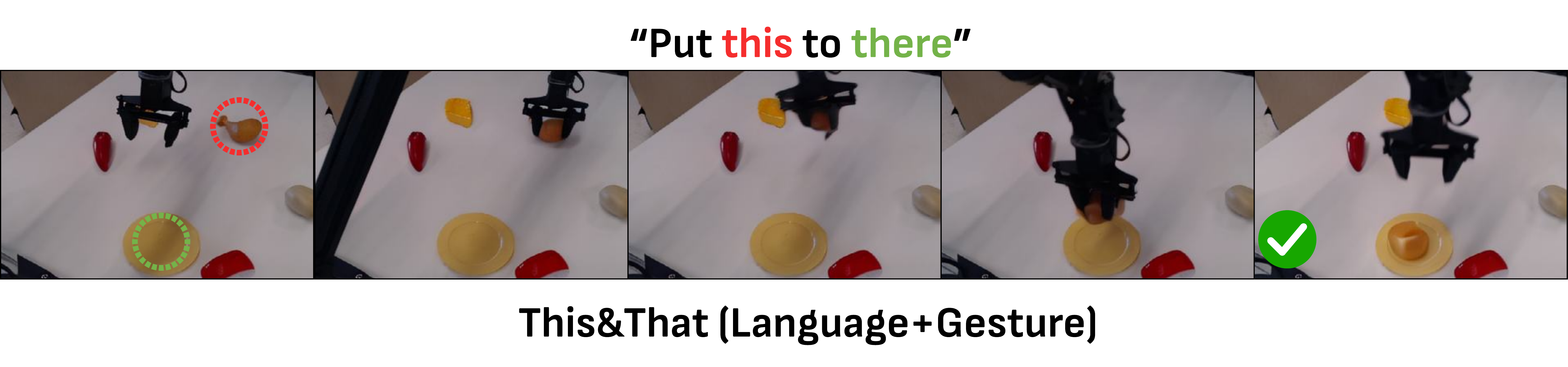} 
    \caption{\textbf{Object Appearance Artifact.} Occasionally, our method generates a video where the object being moved changes its appearance over time, even though the video closely adheres to the user's intention. We believe this issue comes from the inherent limitations of representing 3D geometric relationships within 2D video media.}
    \label{fig:limitation}
    \vspace{5pt}
\end{figure*}

\subsection{Experiment Details}\label{sec:experiment_details}

\subsubsection{VDM Training Details}
In the Bridge dataset training, we use 8 Nvidia L40S GPUs with 48 GB memory each to train 99K iterations (the closest checkpoint to 100K iterations) for the UNet and 4 GPUs with 30K iterations to train the gesture conditioning in the second stage.
The batch size is 1 for each GPU. The pre-trained weight we start with is the default SVD model for the 14-frame version.
For our training on the Isaac Gym dataset, we use 8 GPUs for 30K iterations in the stage 1 and another 4 GPUs for 15K iterations in the stage 2. The pre-trained weight we start with is SVD-XT for the 25-frame version.
We apply the AdamW~\cite{loshchilov2017decoupled} optimizer with a constant learning rate of 1e\mbox{-}5 and 5e\mbox{-}6 respectively for two stages. We apply the 8-bit Adam~\cite{dettmers20218} strategy to decrease GPU memory consumption, and no ema is employed.

To augment the sparsity of the dataset available, we propose a horizontal flip mechanism to augment the dataset. The probability of flipping is 0.45.
However, we keep in mind that if the language prompt contains keywords with position meanings, like \textit{left} and \textit{right}, we will not do flipping.

\subsubsection{VDMs Visual-Quality Experiment Details}
The testing dataset is coming from the train-test split on 10\% of the data from both V1~\cite{ebert2021bridge} and V2~\cite{walke2023bridgedata} to facilitate subsequent quantitative comparisons. For the VDM table, we apply all 646 videos from Bridge V1 after gesture label filtering.

In the VDM quantitative comparisons, we employ FID~\cite{fid}, FVD~\cite{fvd}, PSNR, SSIM, and LPIPS~\cite{zhang2018unreasonable} as our numerical metrics. We compute FID by sampling 9,000 images randomly from the generated frames and the ground truth dataset by the codebase of pyiqa~\cite{pyiqa}. The implementations of FVD, PSNR, SSIM, and LPIPS are from an open source~\footnote{\href{https://github.com/JunyaoHu/common_metrics_on_video_quality}{https://github.com/JunyaoHu/common\_metrics\_on\_video\_quality} }. FVD uses the inception network provided in StyleGAN-V~\cite{skorokhodov2022stylegan}.

All methods first output at the resolution and number of frames available in their codebase and then resize to 256x384 and 14 frames. Specifically, SVD~\cite{blattmann2023stable} outputs at 256x384 with 14 frames and the motion score is 180 with a noise aug strength of 0.1. StreamingT2V~\cite{henschel2024streamingt2v} outputs at 1280x720 with 24 frames and the base model we choose is SVD. We only keep the first 14 frames of 24 frames to compare with GT. DragAnything~\cite{wu2024draganything} outputs at 576x320 with 14 frames. The two drag points we chose are the same gesture points we applied in our gesture conditioning training. We connect two drag points by a straight line. AVDC~\cite{ko2023learning} outputs at 48x64 with 7 frames and 25 sampling steps in the inference stage and the frames are interpolated to 14 frames by repeat. For our model, we directly output at 256x384 with 14 frames for the Bridge testing.

\subsubsection{User Alignment Experiment Details}
For the human study, we selected three individuals with experience in robotics to evaluate our models. Each participant reviewed independently generated results from {\em This\&That} and other baselines to minimize the stochasticity of the study stemming from the generative models' random nature. We selected a total of 24 test cases, divided into specific tasks: 8 cases for pick-and-place, 5 for stacking, 6 for folding, and 5 for opening or closing actions. We compare video outputs of AVDC and our custom VDMs conditioned on {\em first-frame-only, first-frame+language, first-frame+gesture, and first-frame+gesture+langauge (ours)}. For methods that use text inputs, we test both deictic and regular versions.

Given that AVDC is trained using the entire Bridge v1 and v2 datasets, our traditional train-test split method does not fully assess their capabilities. 
Thus, we opt to create new content as conditioning input, which better demonstrates the models' zero-shot generation abilities. This involves altering the language prompts and gesture points to incorporate different objects to pick up and different placement positions. This approach allows us to more effectively evaluate the models' adaptability and performance in novel scenarios.

In our human study, participants are randomly selected common Robotics/CS major bachelor or master students who are not familiar with the purpose of the research and the literature background we have proposed. Participants were presented with an image accompanied by the non-deictic language prompt and gesture points. We then asked them the following question to gauge the alignment of the videos with the ground truth intention:

\textit{``Given one image with the stated intention through a language prompt and gesture points, do you think the generated videos correctly complete the task and align with our intentions?"}

\subsubsection{Preparing Deictic Language Prompts}
The deictic language prompts are generated by automatic scripts that prepend the first action verb to the phrase ``\textit{this to there}", resulting in prompts like ``\textit{put this to there}", ``\textit{fold this to there}", and ``\textit{stack this to there}". These deictic text prompts, despite their ambiguity, are tested for zero-shot capability using the same model weights trained on regular text conditions. Through our experiment, we explore whether simplistic deictic language, such as ``this" and ``there," affects the clarity and alignment of the video generation.

\subsubsection{Training Details of the VDM Baselines}
\begin{itemize}
    \item Vision Model: Utilizes the UNet from Stage 1 designed for the image-to-video task, operating independently of any text prompt.

 \item Vision+Language Model: This version of the UNet, also from Stage 1, incorporates both image and text prompts to generate videos.

 \item Vision+Gesture Model: Employs the temporal ControlNet from Stage 2, using the vision model's pre-trained weights as its backbone.

 \item Vision+Language+Gesture Model: Integrates gesture conditioning training from Stage 2, leveraging the pre-trained UNet weights from the Vision+Language model.
 \item The AVDC model, which inherently accepts image and text prompts as input, was utilized directly with its pre-trained weights for inference to evaluate its effectiveness under the above configurations. 
\end{itemize}

\subsubsection{DiVA Training Details}

We train DiVA using 900 instances for training and 100 instances heldout for testing. Each demonstration consists of approximately 75-100 observation, action pairs. We use a single Nvidia RTX 6000 Ada GPU for 2,000 epochs, which is the same quantity as ACT~\cite{zhao2023learning}. We save checkpoints every 500 epochs and use the checkpoint with the lowest validation error during evaluation. During training, we only use goals sampled from the ground-truth (GT) data. In other words, we subsample the GT visual observations to obtain goals instead of training directly on the generated videos. This is because we retain high performance with such methods. Thus, we can conclude that there is very little domain gap between GT goals and the videos we generate. For our hyperparameters, we use a batch size of 8, a learning rate of $1e^{-5}$, a weight decay of $1e^{-4}$, and an action chunk size of 10. 

\subsubsection{Simulation Experiment Details}

Our dataset consists of pick-and-place demonstrations with four blocks on a tabletop environment. Our blocks are composed of two shapes (cube and cylinder) and eight colors (red, blue, green yellow, magenta, cyan, grey, and black). Demos are generated by relating two randomly selected blocks using five variations: \textit{in front of}, \textit{behind}, \textit{to the right of}, \textit{to the left of}, and \textit{on top of}. We classify the first four relations as \textit{near} tasks and the last as a \textit{stack} task. An example of a language instruction would be \textit{\textbf{stack} the magenta cylinder \textbf{on top of} the black cube}. To show where our method shines, we also create a custom test set with out-of-distribution scenes that contain identical objects in identical colors. In these scenes, we show it is  difficult to use language to precisely define the task and is easier to use gesture instead. 

All demos are collected using a scripted policy. Actions are 7D commands comprising of delta end-effector pose represented in the end-effector frame and a continuous scalar between -1 and 1 to indicate whether to open or close the gripper. During inference (rollout), we execute the full action chunk before querying the model again. Our success metrics are also scripted using a set of basic rules. For pick success, we record the maximum vertical displacement that the object to be picked achieves during rollout. If this value exceeds a block diameter for more than 5 timesteps, we consider the pick to be successful. We divide place success into two separate criteria for near and stack tasks. At each timestep, we calculate the planar and vertical distance from the picked object to its goal location. For near tasks, we count place success if the planar distance to goal is within 3 block diameters and the vertical distance to goal is within a block radius. For stack tasks, we count place success if the planar distance to goal is within a block diameter and the vertical distance to goal is within a block radius. The robot achieves overall success if it has achieved both pick and place success. We rollout actions for 250 timesteps, terminating 5 timesteps after the first instance of overall success. For near tasks, we use more lenient metrics since 2D gestures has some 3D ambiguity. In our experiments, a block diameter of 5cm is used. 

\subsubsection{Baselines in Simulated Experiments}
In Tab.~\ref{table:rollout}, we provide simulation results for four baselines.
In the first row, we train vanilla ACT \cite{zhao2023learning} without any goal-conditioning. For the second row, we condition ACT on language by concatenating the CLIP~\cite{DBLP:journals/corr/abs-2103-00020} embedding of our language prompt with the tokens of the current image observation, where the positional embeddings for CLIP are zero-valued. For the third row, we condition ACT on both language and gesture. The gesture  is encoded with positional embeddings by ResNet in the same way as the current image observation. All the tokens are concatenated together. 
For the 4th row, we retrain AVDC from scratch on demos collected in simulation with low-resolution visual observations. For the 5th row and final baseline, we train our own VDM with only language conditioning to mimic prior works~\cite{yang2023learning, du2023video, zhou2024robodreamer, du2024learning}.
For the 5th row and final baseline, we train our own VDM with only language conditioning to mimic prior works~\cite{yang2023learning, du2023video, zhou2024robodreamer, du2024learning}.

We feed these generated videos to DiVA to generate robot actions. Results on both the regular test set and the out-of-distribution (OOD) test set are provided for all models. As shown in Fig.~\ref{fig:rollout_supp_2} and Fig.~\ref{fig:rollout_supp_1}, the model with Vis.+Lang.+Gesture outperforms all other baselines, especially when the scenes are ambiguous and OOD. 

\subsection{Limitation: Changes in Object Appearance Over Time}\label{sec:limitation_VDM}

While our Video Diffusion Model (VDM) framework typically generates high-quality videos, it occasionally alters the shape of objects being moved over time, as shown in Fig.~\ref{fig:limitation}. We suspect that this issue stems from the inherent limitations of representing 3D geometric relationships within 2D video media. Nevertheless, we believe that these artifacts might not hinder the translation to robotic actions, as long as the visuals of the end-effector remain clear and discernible. 

\clearpage


\end{document}